\documentclass[10pt,twocolumn,letterpaper]{article}

\usepackage{cvpr}              

\usepackage[accsupp]{axessibility}  

\usepackage{graphicx}
\usepackage{amsmath}
\usepackage{amssymb}

\usepackage{tabu}

\usepackage{bm}
\usepackage{pifont}
\newcommand{\cmark}{\ding{51}}%
\newcommand{\xmark}{\ding{55}}%
\usepackage{multirow}

\usepackage{makecell}
\usepackage{color}
\definecolor{Dred}{rgb}{0.6,0,0}
\definecolor{Dgreen}{rgb}{0,0.6,0}

\usepackage[pagebackref,breaklinks,colorlinks]{hyperref}

\usepackage[capitalize]{cleveref}
\crefname{section}{Sec.}{Secs.}
\Crefname{section}{Section}{Sections}
\Crefname{table}{Table}{Tables}
\crefname{table}{Tab.}{Tabs.}

\def\cvprPaperID{5939} 
\def\confName{CVPR}
\def\confYear{2022}

\begin{document}

\title{Revisiting Domain Generalized Stereo Matching Networks from a Feature Consistency Perspective}

\author{Jiawei Zhang\textsuperscript{1}, Xiang Wang\textsuperscript{1}, Xiao Bai\textsuperscript{1}\thanks{Corresponding author: Xiao Bai (baixiao@buaa.edu.cn).}, Chen Wang\textsuperscript{1}, Lei Huang\textsuperscript{2},\\ Yimin Chen\textsuperscript{1},  Lin Gu\textsuperscript{3,4}, Jun Zhou\textsuperscript{5}, Tatsuya Harada\textsuperscript{3,4}, Edwin R. Hancock\textsuperscript{6} \\
\textsuperscript{1}School of Computer Science and Engineering, State Key Laboratory of Software Development \\ Environment, Jiangxi Research Institute, Beihang University, China, \\
\textsuperscript{2}SKLSDE, Institute of Artificial Intelligence, Beihang University, Beijing, China,\\ \textsuperscript{3}RIKEN AIP, Tokyo, Japan, \textsuperscript{4}The University of Tokyo, \textsuperscript{5}Griffith University, \textsuperscript{6}University of York\\
}

\maketitle

\begin{abstract}
Despite recent stereo matching networks achieving impressive performance given sufficient training data, they suffer from domain shifts and generalize poorly to unseen domains. We argue that maintaining feature consistency between matching pixels is a vital factor for promoting the generalization capability of stereo matching networks, which has not been adequately considered. Here we address this issue by proposing a simple pixel-wise contrastive learning across the viewpoints. The stereo contrastive feature loss function explicitly constrains the consistency between learned features of matching pixel pairs which are observations of the same 3D points. A stereo selective whitening loss is further introduced to better preserve the stereo feature consistency across domains, which decorrelates stereo features from stereo viewpoint-specific style information. Counter-intuitively, the generalization of feature consistency between two viewpoints in the same scene translates to the generalization of stereo matching performance to unseen domains. Our method is generic in nature as it can be easily embedded into existing stereo networks and does not require access to the samples in the target domain. When trained on synthetic data and generalized to four real-world testing sets, our method achieves superior performance over several state-of-the-art networks. The code is
available online\footnote{\url{https://github.com/jiaw-z/FCStereo}}.
\end{abstract}

\section{Introduction} \label{sec:intro}

\begin{figure}[t]
\centering
\includegraphics[width=1.0\linewidth]{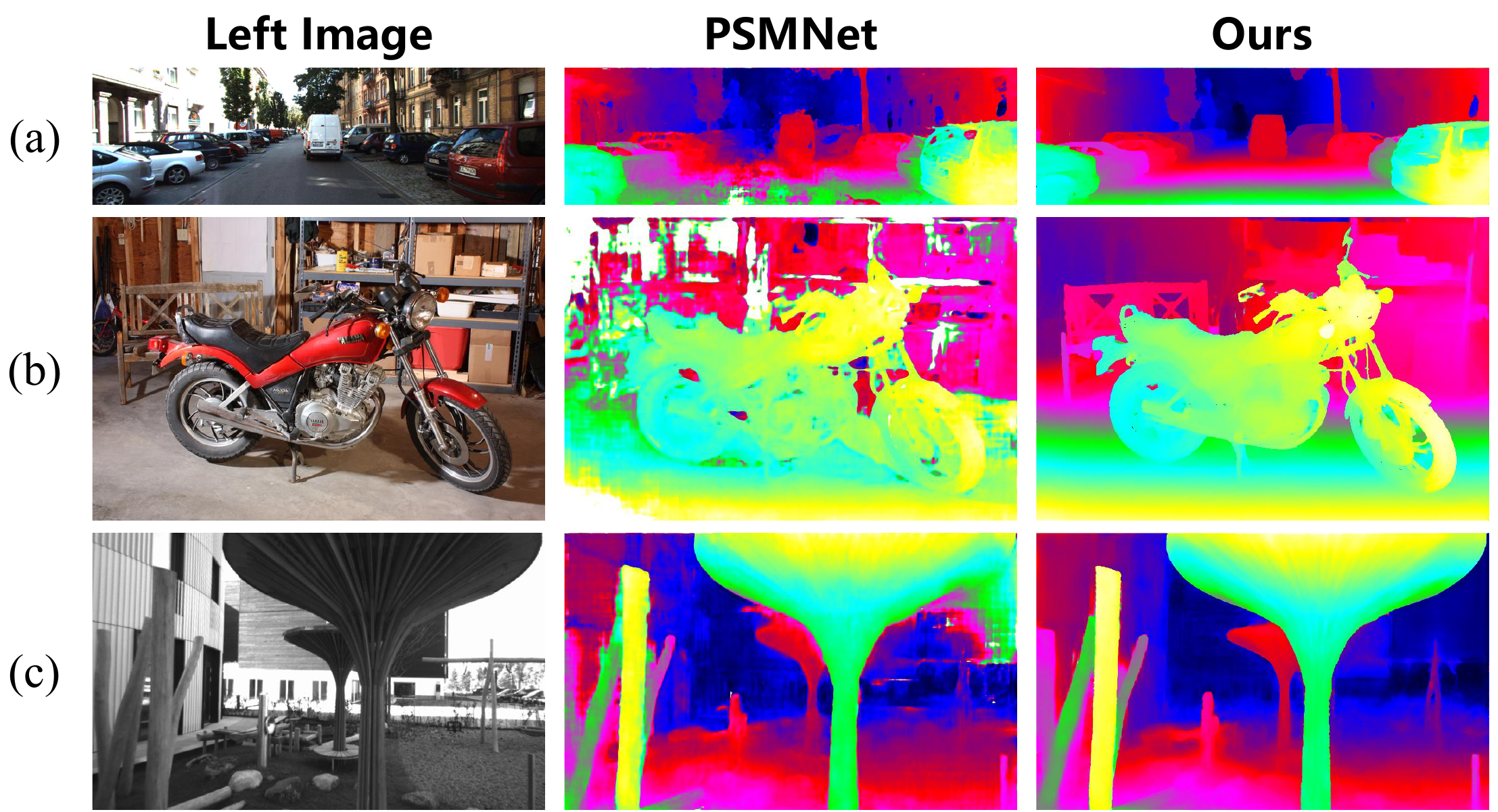}
\vspace{-6mm}
\caption{Domain generalization performance of PSMNet with and without our method on samples from (a) KITTI, (b) Middlebury, and (c) ETH3D training sets. All models are trained on the synthetic SceneFlow dataset.}
\vspace{-3mm}
\label{fig:introduction_show}
\end{figure}

Estimating depth from images is a fundamental problem in many computer vision applications such as autonomous driving \cite{sun2019affordance} and robot navigation \cite{biswas2011depth}. Stereo matching is a solution to this task, which finds the matching correspondences between stereo image pairs and recovers the depth through triangulation. 

Stereo matching is traditionally solved by a matching cost computation process, which usually consists of four steps \cite{scharstein2002taxonomy}: matching cost computation, cost aggregation, disparity regression, and disparity refinement. Recently, end-to-end stereo matching networks \cite{mayer2016large, kendall2017end, chang2018pyramid, guo2019group, zhang2019ga} have been developed based on the cost computation process of traditional methods and achieved state-of-the-art accuracy. However, the poor generalization performance on unseen domains has been a major challenge for their real-world applications (see \Cref{fig:introduction_show} for an example).

A common approach to achieve generalization capability is to learn domain-invariant representations \cite{muandet2013domain, li2018deep, li2018domain, lengyel2021zero}. Some stereo matching networks \cite{zhang2020domain, cai2020matching, shen2021cfnet} have made attempts to tackle this issue by conducting feature-level alignment to obtain domain-invariant features. These works project the inputs into a domain-invariant feature space, reducing the reliance on domain-specific appearance properties and showing more robustness to domain shifts. 

Here, we present a weaker constraint, \textit{stereo feature consistency}, for domain generalized stereo networks. For each point in the left image, stereo matching looks for its matching one in the right view, which naturally requires \textit{robustness to viewpoint changes}. A domain generalized stereo network is expected to generalize this matching ability to unseen domains, which means, in a nutshell, the \textit{generalization of "robustness to viewpoint changes"}. From this perspective, we believe what a stereo network needs to generalize is the matching relationship, behaving as the feature consistency of paired points. For example, traditional methods, which are largely domain-agnostic \cite{tonioni2019unsupervised, poggi2021continual}, compute the matching cost directly on RGB images \cite{hirschmuller2007stereo}. Although image contents differ considerably across different domains, the matching pixels have consistent expressions between the stereo viewpoints in most cases, guaranteeing stable matching cost computation to produce reliable disparity maps. We further verified this intuition to a toy pipeline that combines a cost volume constructed directly from RGB images with the common PSMNet cost aggregation module ($\cf$ \Cref{appendix:rgb cost volume}). Such a simple pipeline with consistent stereo representations also shows a significant improvement in domain generalization performance. 

\begin{figure}[t]
\setlength{\abovecaptionskip}{0.cm}
\setlength{\belowcaptionskip}{-0.cm}
\centering
\begin{subfigure}{1\linewidth}
\centering
\includegraphics[width=0.8\linewidth]{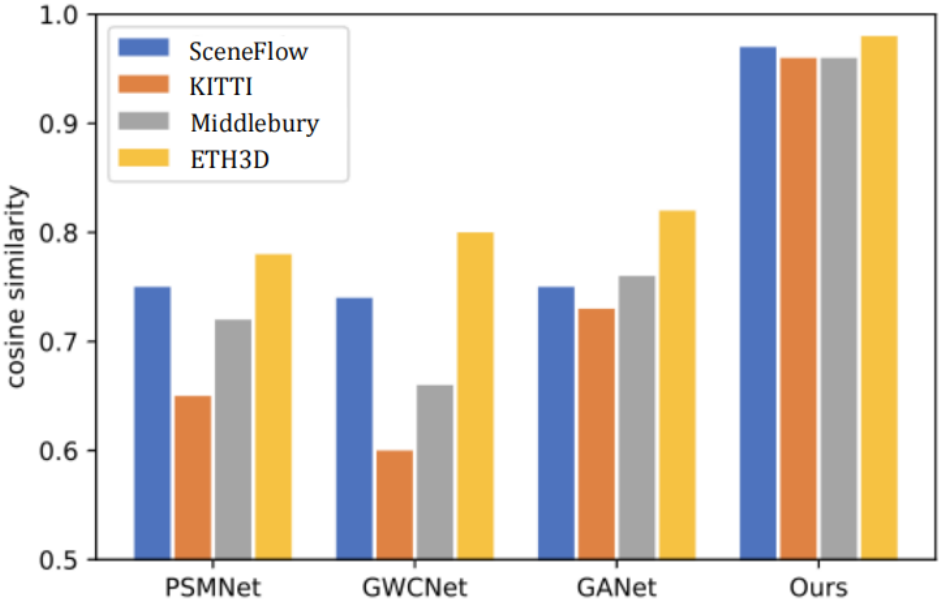}
\caption{}
\vspace{+1mm}
\label{fig:motivation(a)}
\end{subfigure}
\quad
\begin{subfigure}{1\linewidth}
\centering
\includegraphics[width=0.8\linewidth]{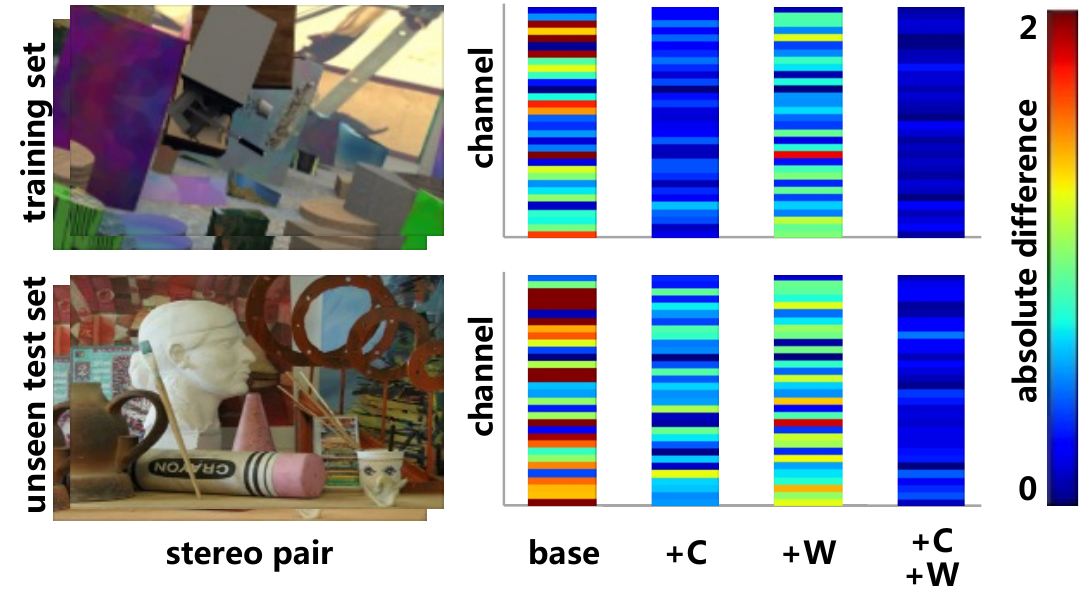}
\vspace{-1mm}
\caption{}
\label{fig:motivation(b)}
\end{subfigure}
\caption{Analysis about feature consistency of matching points. (a) Evaluation of popular stereo matching backbones on four unseen domains. (b) Visualization of per-channel feature inconsistency. Left-right: PSMNet baseline, with our contrastive loss (\textbf{C}), with our whitening loss (\textbf{W}), and with both. More details about learned feature are shown in \Cref{appendix:feature consistency}.}
\vspace{-3mm}
\label{fig:motivation}
\end{figure}

Generally, the appearance inconsistency within a stereo pair is limited to a certain range, thus the matching points being very similar. For example, the corresponding points should share the identical incident light as well as albedo and differ in the shading that appeared in left and right cameras. However, when the learned features are used to construct the cost volume, the feature consistency is not preserved, as shown in \Cref{fig:motivation(a)}. And surprisingly, the features are inconsistent even in the training set, which is contrary to the common intuition that the \textbf{weight-sharing Siamese} feature extractor has dealt with stereo viewpoint changes and extracted consistent features. 

In this paper, we address the domain generalization for stereo matching methods by developing the Feature Consistency Stereo networks (FCStereo). Here comes two challenges: (a) obtaining a high feature consistency on the training set and (b) generalizing this consistency across different domains. We argue that the difficulty of (a) is due to the lack of explicit consistent constraints on features which causes overfitting. We propose the stereo contrastive feature (SCF) loss to encourage the matching points to be close in the representation space. To solve the consistency generalization problem (b), we utilize a proper normalization operation and constrain the feature statistics. A stereo selective whitening (SSW) loss is further introduced to suppress information that is sensitive to stereo viewpoint changes. \Cref{fig:motivation(b)} illustrates the feature differences in a channel-wise manner and shows the role of the two proposed loss terms. SCF loss encourages features to be consistent on the training set. However, we see a degradation of consistency on unseen domains. SSW loss yields a relatively lower consistency compared to the contrastive loss, while the consistency is more robust to domain changes. Jointly using both loss terms enables high feature consistency in various domains. We apply the proposed method to different stereo matching backbones in the experiment and show a significant improvement in generalization performance. It demonstrates that \textbf{the generalization of feature consistency between two viewpoints in the same scene translates to the generalization of stereo matching performance to unseen domains} though appears counter-intuitive. A qualitative illustration is shown in \Cref{fig:introduction_show}. The main contributions of this paper are as follows:
\begin{itemize}
\vspace{-2mm}
\setlength{\topsep}{0pt}
\setlength{\partopsep}{0pt}
\setlength{\itemsep}{3pt}
\setlength{\parsep}{0pt}
\setlength{\parskip}{0pt}
\item We observe that most recent stereo methods learn inconsistent representations for the pairs of matching points, and demonstrate that the generalization performance of stereo networks can be boosted by maintaining a high stereo feature consistency.
\item We propose two loss functions, namely the stereo contrastive feature loss and the stereo selective whitening loss, to encourage the stereo feature consistency across domains. These two losses could be easily embedded in the existing stereo networks.
\item Our method is applied to several stereo network architectures and shows a significant improvement in their domain generalization performance.
\end{itemize}

\section{Related Work}
\textbf{Deep learning based stereo matching.} Since MC-CNN \cite{zbontar2015computing} introduced a convolution neural network (CNN) to matching cost calculation, many deep learning based methods have been proposed for stereo matching. Early works simply replace RGB inputs with expressive learned features for higher accuracy, leaving the following traditional steps for cost computation unchanged. For these methods, a crucial attribute of learned features is the consistency between matching pixels \cite{zhang2017fundamental, fathy2018hierarchical}. 

More recently, many methods solve the task in an end-to-end way \cite{mayer2016large, chang2018pyramid, khamis2018stereonet, zhang2020adaptive, wang2022uncertainty}. Two types of solutions are normally followed by these methods: correlation cost volume based deep neural networks with 2D cost aggregation and concatenation cost volume based stereo networks with 3D cost aggregation. The correlation methods are usually more efficient but cause information loss. DispNetC \cite{mayer2016large} is the first method that introduces end-to-end regression for stereo matching and builds the cost volume in a correlation way. The correlation based matching strategy is adopted by many works \cite{liang2018learning, yin2019hierarchical, tonioni2019real, xu2020aanet} and has achieved impressive and efficient performance. The second category concatenates the stereo features to make full use of information. For example, GCNet \cite{kendall2017end} stacks two view features to build a 4D cost volume and first utilizes 3D convolution for matching cost aggregation. Methods in this category~\cite{chang2018pyramid, wang2019anytime, zhang2019ga,  zhang2020adaptive} leverage more complete information of features and have produced higher accuracy on various stereo benchmarks. Our method can be seamlessly integrated into the existing end-to-end stereo networks and improve their generalization performance. 

\textbf{Domain generalized stereo matching.} It is important to develop stereo matching networks that are robust to unseen domains. DSMNet \cite{zhang2020domain} uses a domain normalization layer to reduce the shifts of image-level styles and local contrast variations, followed by a trainable non-local graph-based filter to smooth the local sensitive local details. CFNet \cite{shen2021cfnet} produces a fused cost volume representation for capturing global and structural information to construct a stereo matching network that is robust to domain changes. Cai $\etal$ \cite{cai2020matching} point out that the poor generalization of stereo networks is caused by the strong dependence of the network on the image appearance, and propose to use a combination of matching functions for feature extraction.

\textbf{Instance discrimination and contrastive learning.}  Instance-level discrimination, regarding each instance as a distinct class of its own, plays an important role in representation learning. This paradigm is formulated as a metric learning problem, where features of positive sample pairs are encouraged to be close and those of negative sample pairs are forced to be apart \cite{wu2018unsupervised}. The following work \cite{spencer2019scale} adopts this idea to specific downstream tasks and shows that the quality of learned representations is heavily affected by the strategy of negative pair selection \cite{spencer2019scale}. Recently, following the idea of instance discrimination, contrastive learning has made remarkable success in self-supervised feature representation learning. MoCo and its variant \cite{he2020momentum, chen2020improved} treat contrastive learning as a dictionary look-up process and maintain a momentum updated queue encoder. Some attempts extend this momentum-based contrastive learning framework to pixel-level feature learning \cite{pinheiro2020unsupervised, xie2021propagate, wang2021dense}. Different from these dense contrastive learning methods, we define positive samples using correspondences given by the ground truth disparity, which is directly tailored to the main task.

\textbf{Feature covariance.} Previous studies have demonstrated that the correlations between feature channels capture the style information of images \cite{gatys2015texture, gatys2016image}. This theory is further explored in style transfer \cite{gatys2016image, li2017universal}, image-to-image translation \cite{cho2019image}, and others \cite{pan2019switchable, luo2017learning}. More recently, \cite{choi2021robustnet} propose a selective whitening method to remove the style information that is sensitive to domain shifts for robust segmentation, where the style information selection depends on the manually designed photometric transformation. Our approach is inspired by the selective whitening \cite{choi2021robustnet}, however, we select the information sensitive to stereo viewpoint changes, without relying on photometric transformation.

\section{Approach}
\begin{figure*}[t]
\centering
\includegraphics[width=1.0\linewidth]{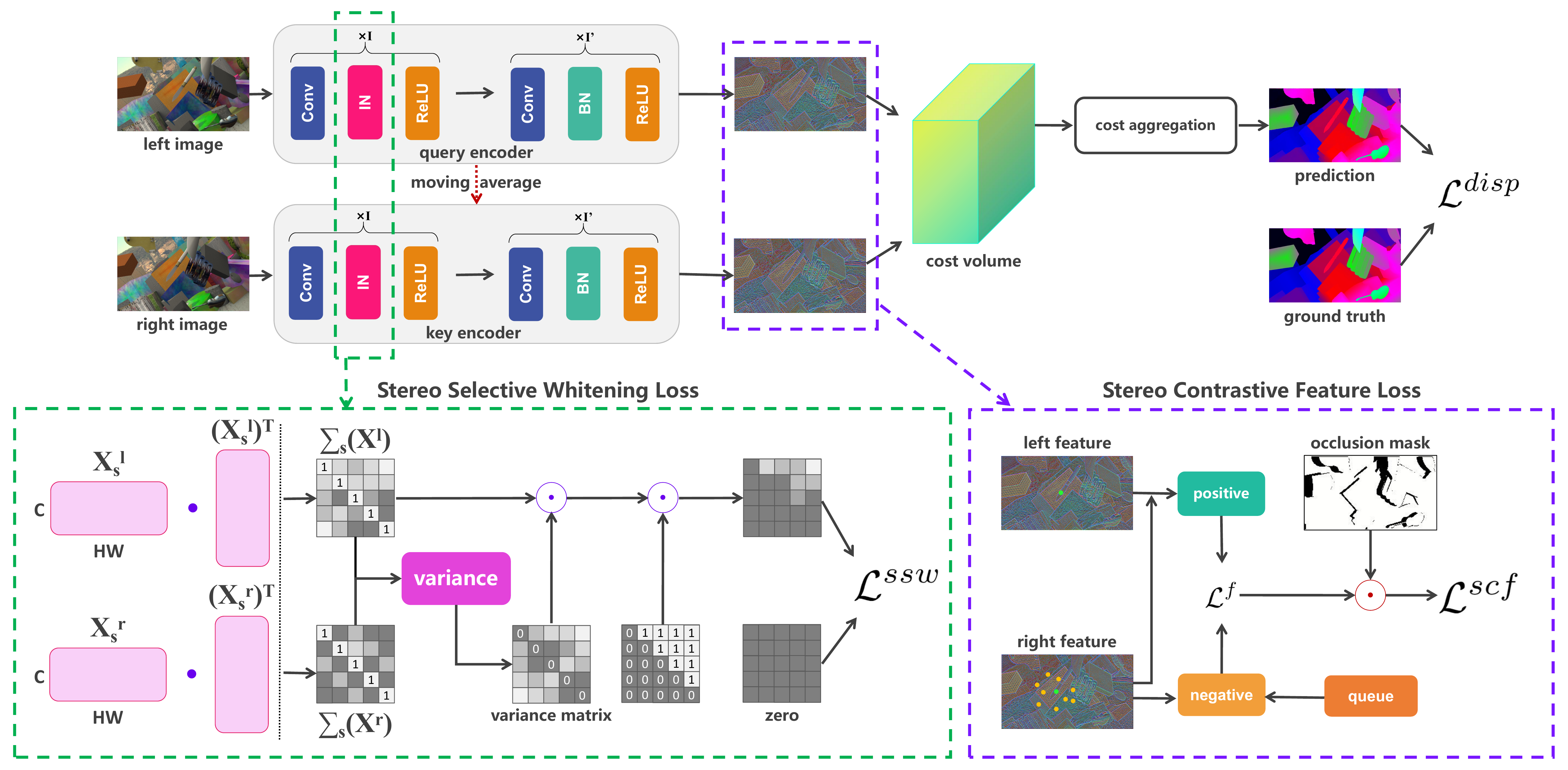}
\vspace{-7mm}
\caption{Structure of method. The top part shows the forward pass that the network extracts features from input pairs and regresses the disparity on the feature based cost volume. At the bottom are the proposed two losses for maintaining the feature consistency. As part of the feature contrastive loss, the key encoder for the right image is implemented as a moving average of the query encoder to alleviate the effects of negative sample selections. In the inference process, we use the query encoder to extract features for both left and right images, which is strictly identical to the standard pipeline.}
\vspace{-2mm}
\label{fig:framework_overview}
\end{figure*}

In this section, we present the details of our method, including a stereo selective whitening loss on the intermediate features and a stereo contrastive feature loss on the final features. \Cref{fig:framework_overview} depicts the whole framework of our method, where the stereo contrastive feature loss and stereo selective whitening loss are applied to a stereo matching network for encouraging the stereo feature consistency across domains.

\subsection{Stereo Contrastive Feature Loss}
Stereo features from the last feature extraction layer are used to construct the cost volume, which is the most important internal representation in a deep stereo network. At this stage, we impose a consistency constraint on the features of stereo views. Inspired by the recent success of contrastive learning on unsupervised feature learning via optimizing the pairwise (dis)similarity, we introduce a contrastive learning mechanism on stereo features, namely the Stereo Contrastive Feature (SCF) loss. The proposed contrastive learning mechanism includes a pixel-level contrastive loss applied on stereo features, and a dictionary queue with a momentum updated key encoder, which introduces a rich set of negative samples from different pairs and further improves the feature consistency.

\textbf{Positive pairs.} We consider the pixel vectors in stereo views as a positive pair if their pixel coordinates are the projected locations of the same 3D point. These positive pairs can be collected using the ground truth disparity $d$ of the left view, $\ie$, the query feature in the left view $\boldsymbol{\phi}^{l}_{u,v}$ are paired with the key feature in the right view $\boldsymbol{\phi}^{r}_{u-d,v}$. Therefore, maintaining \textit{stereo feature consistency} simply becomes promoting the feature consistency of positive pairs.

\textbf{Negative pairs.} While each of the pixels in $\boldsymbol{\phi}^l$ has $HW-1$ potential negative pairs in $\boldsymbol{\phi}^r$, involving all potential negative pairs in the contrastive loss would lead to a huge computational cost. To overcome this issue, we use a naive method that randomly samples $N$ non-matching points from the right feature $\boldsymbol{\phi}^{r}$ in a local window with a size of $50\times 50$ to form $N$ negative pairs.

\textbf{Momentum encoder.}
The selection strategy of negative samples is particularly important as it heavily affects the property of learned representations \cite{spencer2019scale, xie2021propagate}. In our experiment, the feature consistency is not high enough when the negative samples are limited to the same stereo pair. Using pixels from other images as negative samples can be more natural \cite{pinheiro2020unsupervised} and it fits well with a dictionary queue for negative samples. 

We follow \cite{he2020momentum} to maintain a dynamic dictionary queue that stores preceding negative samples and change the architectural design of weight-sharing feature extractors into an asymmetric pair of query and key encoders (with weights $\boldsymbol{\theta}$ and $\boldsymbol{\eta}$). The capacity of the queue is fixed as $K$ and the oldest samples are progressively replaced after each iteration. The key encoder is modeled as a momentum-based moving average of the query encoder:
\vspace{-2mm}
\begin{equation}
    \boldsymbol{\eta}_{t} = m\boldsymbol{\eta}_{t-1} + (1-m)\boldsymbol{\theta}_{t},
\label{eq:momentum}
\vspace{-1mm}
\end{equation}
where $t$ is the number of iterations and $m \in [0,1]$ is a momentum value. Such a design can provide content from different images for negative samples, which reduces the opportunity for features to focus too much on the contents of the current image. As the momentum value $m$, plays a core role in making use of a queue \cite{he2020momentum}, we evaluate the feature consistency in the experiment. It indicates that a relatively large momentum value ($\eg, m=0.9999$, our default) plays a core role in extracting stereo consistent representations.

\textbf{Pixel-wise contrastive loss.} We measure the similarity of feature pairs with dot product, and adopt a pixel-level InfoNCE~\cite{pathak2017learning} to our problem:
\begin{equation}
    \mathcal{L}^{f}(u,v) = -\log \frac{
    \exp{
    (\boldsymbol{\phi}^{l}_{u,v} 
    \cdot 
    \boldsymbol{\phi}^{r}_{u-d,v}
    / \tau)}}
    {\sum_{\boldsymbol{\phi}^n\in\mathcal{F}(u,v)}\exp{
    (\boldsymbol{\phi}^{l}_{u,v}
    \cdot 
    \boldsymbol{\phi}^{n}
    / \tau)}},
\label{contrastive_loss}
\end{equation}
where $\mathcal{F}(u,v)$ denotes the negative sample set of sample $\boldsymbol{\phi}^l_{u,v}$, consisting $N$ samples from the right feature $\boldsymbol{\phi}^r$ and $K$ samples from the dictionary queue, and $\tau$ is a temperature hyper-parameter~\cite{wu2018unsupervised}. We set $N=60$, $K=6000$, and $\tau=0.07$.

\textbf{Non-matching region removal.}
We leverage ground truth disparity to collect pixel feature pairs as positive pairs in the contrastive framework. However, some pairs collected in this way don't originate from the same point in the 3D world, due to factors like occlusions. Hence these \textit{non-matching} pairs should be detected and eliminated from the positive sample set. A widely used matching confidence criterion, left-right geometric consistency check, can be leveraged to detect and remove those non-matching sample pairs. The reprojection error $\mathbf{R}$ is computed as the difference of ground truth disparity values at paired pixel locations in stereo images and could be served as a criterion for matching validity check. Then the mask $\mathbf{M}$ denoting the remaining matching regions is defined as:
\begin{equation}
\begin{split}
\mathbf{M}_{u,v} = \left \{
\begin{array}{ll}
1, &\mathbf{R}_{u,v} < \delta\\
0, &otherwise \\
\end{array}
\right.
\end{split}
\end{equation}
where $\delta$ is set to 3 as a threshold. And our SCF loss is defined as the weighted average of $\mathcal{L}^{f}$ over the pixel coordinate space $\mathcal{C}$:
\begin{equation}
    \mathcal{L}^{scf} = \frac{1}{\sum_{(u,v)\in\mathcal{C}}\textbf{M}_{u,v}}\sum_{(u,v)\in\mathcal{C}}L^{f}(u,v) \odot \textbf{M}_{u,v}.
\end{equation}

\subsection{Stereo Selective Whitening Loss}

With the contrastive loss, stereo network extracts consistent representation on the training set. However, the degradation of feature consistency across different domains has become the main obstacle to the further improvement of generalization performance. We build the stereo selective whitening (SSW) loss based on \cite{choi2021robustnet} to address this problem.

Generally, stereo networks use batch normalization (BN) \cite{ioffe2015batch} as their default feature normalization operation. During training, BN regularizes the feature with the mini-batch statistics and uses population statistics of the training set during inference \cite{huang2017arbitrary}, which makes the statistics of networks data-dependent \cite{zhang2020domain} and is sensitive to the change of domains. To generalize the feature consistency to different domains, we change some default BN layers into instance normalization (IN) \cite{ulyanov2016instance} layers, which regularizes each sample separately, therefore, is independent of training set statistics. For each sample $\mathbf{X}\in \mathbb{R}^{C \times HW}$, IN transforms it into $\hat{\mathbf{X}}\in \mathbb{R}^{C \times HW}$ :
\begin{equation}
    \hat{\mathbf{X}}_{i} =  \frac{1}{\mathbf{\sigma}_{i}} (\mathbf{X}_{i} - \mathbf{\mu}_{i}),
\end{equation}
where $\mathbf{\mu}_{i}$ and $\mathbf{\sigma}_{i}$ are mean and the standard deviation of $\hat{\mathbf{X}}$ along the channel index $i$.

We further consider the information stored in the feature covariance, which is not dealt with by IN. The proposed SSW seeks for learning viewpoint-invariant representation by suppressing the feature covariance components that are sensitive to stereo view changes. In particular, we firstly compute the variance matrix $\mathbf{\Sigma}(\hat{\mathbf{X}}) \in \mathbb{R}^{C \times C}$ of the IN regularized representation $\hat{\mathbf{X}}$:
\begin{equation}
    \mathbf{\Sigma}(\hat{\mathbf{X}}) = \frac{1}{HW} (\hat{\mathbf{X}}) (\hat{\mathbf{X}})^{\mathrm{T}}.
\end{equation}
We then compute the covariance matrices $\mathbf{V} \in \mathbb{R}^{C \times C}$ between each left view feature covariance $\mathbf{\Sigma}(\hat{\mathbf{X}}^{l})$ and its corresponding right feature $\mathbf{\Sigma}(\hat{\mathbf{X}}^{r})$, where $n$ indexes the samples:
\begin{equation}
\begin{split}
    \boldsymbol{\mu}_{\mathbf{\mathbf{\Sigma}}_{n}} &= \frac{1}{2}(\mathbf{\Sigma}_{n}(\hat{\mathbf{X}}^{l}) + \mathbf{\Sigma}_{n}(\hat{\mathbf{X}}^{r})) \\
    \mathbf{V} &= \frac{1}{2N} \sum^{N}_{n=1} ((\mathbf{\Sigma}_{n}(\hat{\mathbf{X}}^{l}) - \boldsymbol{\mu}_{\mathbf{\Sigma}_{n}})^{2} + (\mathbf{\Sigma}_{n}(\hat{\mathbf{X}}^{r}) - \boldsymbol{\mu}_{\mathbf{\Sigma}_{n}})^{2}),
\end{split}
\end{equation}
The element $\mathbf{V}_{i,j}$ from the variance matrix measures how sensitive the correspondence between the $i$-th and the $j$-th channels to stereo viewpoint changes. Covariance elements with high variances between left and right features, which are considered as sensitive components to stereo view changes, should be considered in the whitening loss. In practice, all covariance elements are grouped into 3 clusters by the magnitudes of variance \cite{choi2021robustnet}, and we choose the one with the highest variance value, termed $\mathcal{G}_p$. Then a selective mask $\tilde{\mathbf{M}} \in \mathbb{R}^{C \times C}$ is computed:
\begin{equation}
\begin{split}
\tilde{\mathbf{M}}_{i,j}= \left \{
\begin{array}{ll}
1, &\mathbf{V}_{i,j} \in \mathcal{G}_p\\
0, &otherwise \\
\end{array}
\right.
\end{split}
\end{equation}
The SSW loss is imposed on the left regularized features:
\begin{equation}
    \mathcal{L}^{ssw} = \frac{1}{\Gamma} \sum_{\gamma=1}^{\Gamma} ||\mathbf{\Sigma}_{\gamma}(\hat{\mathbf{X}}^{l}) \odot \tilde{\mathbf{M}} \odot \hat{\mathbf{M}}||_{1},
\end{equation}
where $\hat{\mathbf{M}}$ is a strict upper triangular matrix as the covariance matrix is symmetric, $\Gamma$ is the number of layers to which the SSW loss is applied, and $\gamma$ indexes the corresponding layer (\ie conv1, conv2\_x in PSMNet). With the SSW loss, the stereo network learns to rely less on stereo irrelevant information to form its feature representation. The differences within a stereo image pair are mostly limited to specific physical characteristics, such as diffuse reflection of light, which gives the stereo model the possibility to learn some general knowledge from limited training data.

\begin{table*}[t]
\begin{center}
\begin{tabular}{c|ccc|ccccc}
\hline
\multirow{2}{*}{Backbone} & Contrastive  & Momentum & Stereo & \multicolumn{2}{c}{KITTI} & \multicolumn{2}{c}{Middlebury} & \multirow{2}{*}{ETH3D} \\
& Loss  & Encoder & Whitening & 2012 & 2015 & half & quarter & \\
\hline\hline
\multirow{5}{*}{PSMNet \cite{chang2018pyramid}}
& \textcolor{Dred}{\xmark} & \textcolor{Dred}{\xmark} & \textcolor{Dred}{\xmark} & 26.5 & 27.9 & 26.9 & 20.0 & 23.8 \\
& \textcolor{Dgreen}{\cmark} & \textcolor{Dred}{\xmark} & \textcolor{Dred}{\xmark} & 18.4 & 19.0 & 24.1 & 15.4 & 17.6 \\
& \textcolor{Dgreen}{\cmark} & \textcolor{Dgreen}{\cmark} & \textcolor{Dred}{\xmark} & 10.5 & 12.7 & 22.2 & 15.0 & 17.1 \\
& \textcolor{Dred}{\xmark} & \textcolor{Dred}{\xmark} & \textcolor{Dgreen}{\cmark} & 13.2  & 15.5 & 20.5 & 13.8  & 14.1 \\
& \textcolor{Dgreen}{\cmark} & \textcolor{Dgreen}{\cmark} & \textcolor{Dgreen}{\cmark} & \textbf{7.0}  & \textbf{7.5} & \textbf{18.3}  & \textbf{12.1} & \textbf{12.8} \\
\hline\hline
\multirow{4}{*}{GWCNet \cite{guo2019group}}
& \textcolor{Dred}{\xmark} & \textcolor{Dred}{\xmark} & \textcolor{Dred}{\xmark} & 20.2 & 22.7 & 34.2 & 18.1 & 30.1 \\
& \textcolor{Dgreen}{\cmark} & \textcolor{Dred}{\xmark} & \textcolor{Dred}{\xmark} & 12.3 & 16.5 & 25.8 & 15.5 & 13.3 \\
& \textcolor{Dgreen}{\cmark} & \textcolor{Dgreen}{\cmark} & \textcolor{Dred}{\xmark} & 11.2 & 12.1 & 24.8 & 15.2 & 12.8\\
& \textcolor{Dred}{\xmark} & \textcolor{Dred}{\xmark} & \textcolor{Dgreen}{\cmark} & 12.0 & 13.5 & 24.6  & 14.9  & 12.5 \\
& \textcolor{Dgreen}{\cmark} & \textcolor{Dgreen}{\cmark} & \textcolor{Dgreen}{\cmark} & \textbf{7.4} & \textbf{8.0}  & \textbf{21.0} & \textbf{11.8}  & \textbf{11.7}  \\
\hline\hline
\multirow{4}{*}{GANet \cite{zhang2019ga}}
& \textcolor{Dred}{\xmark} & \textcolor{Dred}{\xmark} & \textcolor{Dred}{\xmark} & 10.1 & 11.7 & 20.3 & 11.2 & 14.1 \\
& \textcolor{Dgreen}{\cmark} & \textcolor{Dred}{\xmark} & \textcolor{Dred}{\xmark} & 9.1 & 9.5 & 18.1 & 10.5 & 12.1 \\
& \textcolor{Dgreen}{\cmark} & \textcolor{Dgreen}{\cmark} & \textcolor{Dred}{\xmark} & 7.2 & 7.5 & 16.3 & 10.1 & 11.3 \\
& \textcolor{Dred}{\xmark} & \textcolor{Dred}{\xmark} & \textcolor{Dgreen}{\cmark} & 8.4  & 9.0 & 16.8  & 10.2 & 10.5 \\
& \textcolor{Dgreen}{\cmark} & \textcolor{Dgreen}{\cmark} & \textcolor{Dgreen}{\cmark} & \textbf{5.7}  & \textbf{6.4} & \textbf{16.0}  & \textbf{9.8}  & \textbf{9.2}  \\
\hline
\end{tabular}
\end{center}
\vspace{-6mm}
\caption{Ablation study of each key component with various backbones on the KITTI, Middlebury, and ETH3D training sets. Threshold error rates (\%) are adopted for evaluation.}
\vspace{-1mm}
\label{table:exp1:ablation_study} 
\end{table*}

\subsection{Training Objectives}
Our final training loss is a weighted sum of the disparity loss and above-mentioned losses:
\begin{equation}
    \mathcal{L} = \mathcal{L}^{disp} + \lambda^{scf}\mathcal{L}^{scf} + \lambda^{ssw}\mathcal{L}^{ssw},
\end{equation}
where $\mathcal{L}^{disp}$ is a commonly used per-pixel smooth-$\mathcal{L}_1$ loss for disparity regression. $\lambda^{scf}$ and $\lambda^{ssw}$ are the balancing weights. During back-propagation, all other modules are updated by classical gradient descent methods, except that the right feature extractor is implemented as a moving average of the left extractor.

\vspace{-0.1in}
\section{Experiments}
In this section, we make a detailed analysis of some commonly used stereo methods to illustrate that the existing framework lacks explicit constraints on features. We also perform ablation studies on different datasets, including KITTI \cite{geiger2012we, menze2015object}, Middlebury \cite{scharstein2014high}, ETH3D \cite{schops2017multi}, DrivingStereo \cite{yang2019drivingstereo}, to verify the role of different components. We compare our method with existing domain generalized stereo networks to show the effectiveness of our method.

\subsection{Datasets}
\noindent \textbf{SceneFlow} \cite{mayer2016large} is a large synthetic dataset containing three subsets: Driving, Monkaa, and FlyingThings3D. The training set includes $35k$ pairs of synthetic stereo images and dense ground-truth disparities with a resolution of $960\times540$, which is used to train networks from scratch in our experiments.

\noindent \textbf{KITTI2012} \cite{geiger2012we} and \textbf{KITTI2015} \cite{menze2015object} both collect outdoor driving scenes with a full resolution of $1242\times375$. They provide 394 stereo pairs with sparse ground-truth disparities for training and 395 pairs for testing. We use the training sets to evaluate the generalization performance of networks.

\noindent \textbf{Middlebury 2014} \cite{scharstein2014high} is an indoor dataset, providing 28 training (including 13 additional stereo pairs) and 15 testing stereo pairs with full, half, and quarter resolutions. We use half and quarter resolution training sets to evaluate the generalization ability of networks.

\noindent \textbf{ETH3D} \cite{schops2017multi} consists of 27 grayscale image pairs for training and 20 for testing. It includes both indoor and outdoor scenes. We use the training set for generalization performance evaluation.

\noindent \textbf{DrivingStereo} \cite{yang2019drivingstereo} is a large-scale real dataset. A subset of it contains 2,000 stereo pairs collected under different weathers (sunny, cloudy, foggy, and rainy). We evaluate the generalization performance on these challenging scenes.

\subsection{Implementation Details}
We implement our method in PyTorch and train it with Adam optimizer ($\beta_{1}=0.9$, $\beta_{2}=0.999$). The batch size is set to 12 on GPUs. We train the models from scratch with the learning rate of 0.001 for 15 epochs and 0.0001 for further 5 epochs. We randomly crop the raw images to $512\times256$ as input. For all datasets, color normalization is used with the mean ([0.485, 0.456, 0.406]) and variation ([0.229, 0.224, 0.225]) of the ImageNet \cite{deng2009imagenet} for data pre-processing. We set the maximum disparity as $D=192$, and all ground-truth disparities larger than $D$ are excluded from the loss calculation. During training, we use asymmetric query and key encoders to extract features from left and right images, respectively. In the test phase, the query encoder is used as the feature extractor for both left and right images, which is the symmetric design strictly identical to the standard stereo pipeline.

\subsection{Ablation Study}
In this section, We present detailed ablation studies to evaluate and analyze the effectiveness of our method.

\noindent \textbf{Key components:} 
We evaluate the effectiveness of each key component of our pipeline. Here, three networks are selected as baseline models. PSMNet \cite{chang2018pyramid} is a widely-adopted backbone. It constructs a concatenation based cost volume and hopes that the cost aggregation network can learn a similarity measurement function from scratch. GWCNet \cite{guo2019group} is selected as it constructs the cost volume with a group-wise correlation, which provides better similarity measures than learning from scratch. GANet \cite{zhang2019ga} is one of the top-performing networks, guiding the cost aggregation with low-level features. As shown in \Cref{table:exp1:ablation_study}, applying the contrastive loss on the final features during training significantly improves the domain generalization performance. Benefiting from the negative samples from different stereo pairs and the momentum encoder, it generalizes better to various domains, $\eg$ $15.2\%$ on KITTI, $4.7\%$ on Middlebury, and $6.7\%$ on ETH3D for PSMNet. In addition, the error rates on unseen domains are reduced by the whitening loss. It shows that maintaining the feature consistency across different domains can effectively improve the generalization performance. Furthermore, with the combination of these key components, the models significantly outperform their corresponding baseline models in unseen domains.

\noindent \textbf{Momentum value:} \Cref{table:momentum_value} shows the consistency of learned stereo features with different momentum values ($m$ in \Cref{eq:momentum}). Compared to the standard Siamese encoder without the dictionary queue, our momentum encoder is shown beneficial for the feature consistency of positive pairs, and this behavior holds for both seen and unseen domains. We also adopt different momentum values $m$ and show that a relatively large $m$ is vital to achieve higher stereo feature consistency. 

\begin{table}[t]\footnotesize
\begin{center}
\begin{tabu}{c|cccccc}
\emph{$m$} & - & 0.9 & 0.99 & 0.999 & 0.9999 & 0.99999 \\
\tabucline[1.2pt]{-}
SceneFlow & 0.86 & 0.88 & 0.92 & 0.96 & 0.97 & 0.98\\
KITTI & 0.78 & 0.82 & 0.85 & 0.91 & 0.92 & 0.92
\end{tabu}
\end{center}
\vspace{-6mm}
\caption{Feature consistency with different momentum values in both seen (SceneFlow) and unseen (KITTI) domains. Cosine similarity is adopted as the vector-wise similarity metric for evaluation. '-' denotes the contrastive learning setting without queue and momentum encoder. We select PSMNet as the baseline model.}
\vspace{-2mm}
\label{table:momentum_value} 
\end{table}

\begin{table}[t]\footnotesize
\begin{center}
\begin{tabular}{c|c|ccc}
\hline
Methods &
KITTI & Middlebury \\
\hline\hline
PSMNet Baseline  & 12.7 & 22.2 \\
+ Instance Norm \cite{ulyanov2016instance} & 8.5 & 19.1  \\
+ Domain Norm \cite{zhang2020domain} & 8.1 & 18.8  \\
+ Instance Whitening \cite{choi2021robustnet} & 8.0  & 18.6   \\
+ Our Stereo Selective Whitening  & \textbf{7.5} & \textbf{18.3} \\
\hline
\end{tabular}
\end{center}
\vspace{-5mm}
\caption{Comparisons with existing normalization layers on the KITTI 2015 and the half resolution Middlebury training sets. Threshold error rates (\%) are adopted. We select PSMNet as the baseline model.}
\vspace{-4mm}
\label{table:stereo_whitening} 
\end{table}

\noindent \textbf{Normalization layer:} We evaluate the effectiveness of our proposed stereo whitening with batch normalization \cite{ioffe2015batch}, instance normalization \cite{ulyanov2016instance}, domain normalization \cite{zhang2020domain} and instance whitening \cite{choi2021robustnet}. All other settings except for the normalization method are kept the same in the experiment. Compared with the general methods, our whitening loss is specifically designed for stereo matching and helps the model to generalize better to unseen domains, as shown in \Cref{table:stereo_whitening}.

\noindent \textbf{Feature consistency of various architectures:} We present the feature consistency and the generalization performance of various networks. We use the cosine similarity as a vector-wise metric to evaluate the consistency between matching features. The aforementioned baseline methods, PSMNet \cite{chang2018pyramid}, GWCNet \cite{guo2019group}, GANet \cite{zhang2019ga}, are included in this experiment. AANet \cite{xu2020aanet} is selected as the representation for full correlation cost volume based methods \cite{mayer2016large, xu2020aanet}. We also evaluate the CasPSMNet \cite{gu2020cascade} as the representation for coarse-to-fine methods. DSMNet \cite{zhang2020domain} is designed for domain generalization and extracts the non-local features. In addition to the common disparity regression loss, AcfNet \cite{zhang2020adaptive} imposes additional constraints on the probability distribution derived from the filtered cost volume. CDN \cite{garg2020wasserstein} replaces the commonly used softargmin-based regression loss with a Wasserstein distance based loss. As shown in \Cref{table:exp:frameworks_network}, these popular methods all lack explicit constraints on features and extract inconsistent representations.

\begin{table}[t]\footnotesize
\begin{center}
\begin{tabular}{c|cc|cc}
\hline
\multirow{2}{*}{Methods} &\multicolumn{2}{c|}{KITTI} & \multicolumn{2}{c}{Middlebury} \\ \cline{2-5}
& cosine & \textgreater3px  & cosine & \textgreater2px  \\
\hline\hline
AANet \cite{xu2020aanet} & 0.76 & 12.3  & 0.77 & 28.1 \\
Cas-PSMNet \cite{gu2020cascade} & 0.58 & 16.5 & 0.63 & 27.8 \\
AcfNet \cite{zhang2020adaptive} & 0.58 & 27.4  & 0.61  & 27.1  \\
CDN-PSMNet \cite{garg2020wasserstein} & 0.57 & 40.0 & 0.61 & 35.0  \\
PSMNet \cite{chang2018pyramid} & 0.65 & 27.9 & 0.71 & 26.9  \\
GWCNet \cite{guo2019group} & 0.60 & 22.7  & 0.67 & 34.2 \\
GANet \cite{zhang2019ga} & 0.73 & 11.7  & 0.76 & 20.3  \\
DSMNet \cite{zhang2020domain} & 0.83 & 6.5 & 0.85 & 13.8 \\
\hline
FC-PSMNet (ours)  & 0.98 & 7.5 & 0.95 & 18.3 \\
FC-GWCNet (ours) & 0.97 & 8.0 & 0.95 & 21.0 \\
FC-GANet (ours) & 0.98 & 6.4 & 0.97 & 16.0 \\
FC-DSMNet (ours) & \textbf{0.99} & \textbf{6.2} & \textbf{0.98} & \textbf{12.0} \\
\hline
\end{tabular}
\end{center}
\vspace{-5mm}
\caption{Evaluation of feature consistency and generalization performance on the KITTI 2015 and the half resolution Middlebury training sets. Cosine similarity and threshold error rates (\%) are adopted.}
\vspace{-2mm}
\label{table:exp:frameworks_network} 
\end{table}

\subsection{Cross-domain Evaluation}
We compare our methods with several other stereo matching methods, including traditional methods, well-researched end-to-end methods, and domain generalized methods by training on synthetic SceneFlow training set and evaluating on four real-world datasets. The asymmetric augmentation \cite{yang2019hierarchical, watson2020learning} is used to prevent the model from overfitting. 
\Cref{table:exp:cross-domain_evaluations} summarizes the comparisons. Our method achieves superior generalization performance than others.

\subsection{Evaluation on Challenging Weathers}
\begin{figure}[t]
\centering
\includegraphics[width=1.0\linewidth]{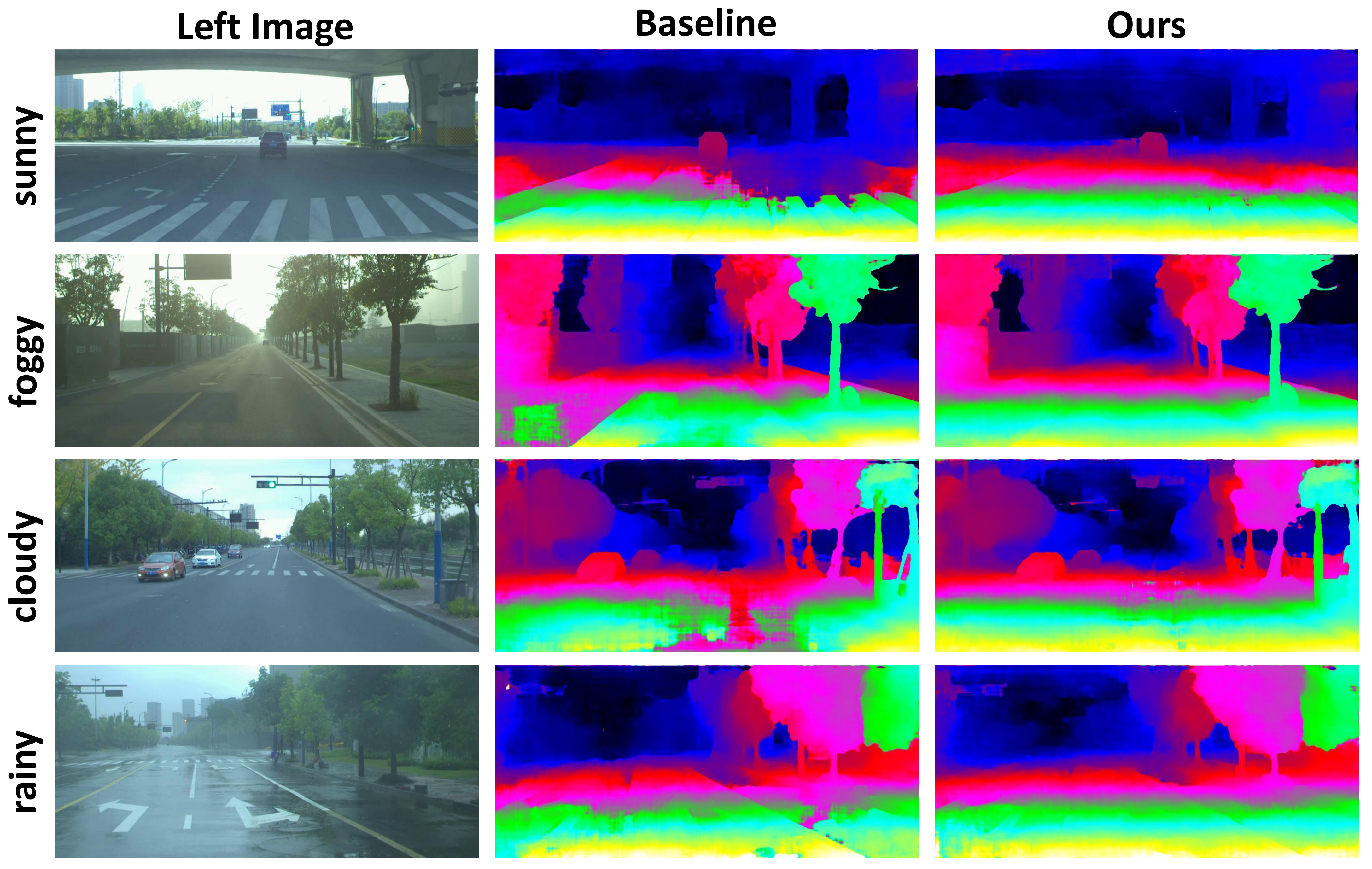}
\vspace{-6mm}
\caption{Qualitative results on different weather conditions of DrivingStereo. PSMNet is selected as our baseline model.}
\vspace{-2mm}
\label{fig:weather}
\end{figure}

In this section, we evaluate the generalization performance of our method on some challenging domains. We train the baseline models and our models under the asymmetric augmentation. The trained models are tested on stereo pairs collected in four challenging weather conditions provided by DrivingStereo. The KITTI 2015 training set is also evaluated as it collects similar outdoor driving scenes under ideal weather conditions. The results are summarized in \Cref{table:exp:weather}. Compared with baseline models, our models generalize better to images under ideal weather conditions, and the improvement is more obvious in challenging weather conditions. \Cref{fig:weather} shows the qualitative results.

\begin{table}[t]\footnotesize
\begin{center}
\begin{tabular}{c|ccccc}
\hline
\multirow{2}{*}{Methods} & \multicolumn{2}{c}{KITTI} & \multicolumn{2}{c}{Middlebury} & \multirow{2}{*}{ETH3D} \\
& 2012 & 2015 & half & quarter & \\
\hline\hline
CostFilter~\cite{hosni2012fast} & 21.7 & 18.9 & 40.5 &  17.6 & 31.1 \\
PatchMatch~\cite{bleyer2011patchmatch} & 20.1 &  17.2 &  38.6 & 16.1 & 24.1 \\
SGM~\cite{hirschmuller2007stereo} & 7.1 &  7.6 &  25.2 &  10.7 & 12.9 \\
\hline
Training set & \multicolumn{5}{c}{SceneFlow} \\
\hline
PSMNet~\cite{chang2018pyramid} & 26.5 & 27.9 & 26.9 & 20.0 & 23.8 \\
GWCNet~\cite{guo2019group} & 20.2 & 22.7 & 34.2 & 18.1 & 30.1 \\
GANet~\cite{zhang2019ga} & 10.1 & 11.7 & 20.3 & 11.2 & 14.1 \\
MS-PSMNet \cite{cai2020matching} & 13.9 & 7.8 & 19.9 & 10.8  & 16.8 \\
MS-GCNet \cite{cai2020matching} & \textbf{5.5} & 6.2 & 18.5 & 10.3  & 8.8 \\
DSMNet~\cite{zhang2020domain} & 6.2 & 6.5 & 13.8 & 8.1 & 6.2 \\
FC-PSMNet (ours) & 7.0 & 7.5 & 18.3 & 12.1  & 12.8 \\
FC-GWCNet (ours) & 7.4 & 8.0 & 21.0 & 11.8  & 11.7  \\
FC-GANet (ours) & 5.7 & 6.4 & 16.0 & 9.8  & 9.2 \\
FC-DSMNet (ours) & \textbf{5.5} & \textbf{6.2}  & \textbf{12.0}  & \textbf{7.8}  & \textbf{6.0}  \\
\hline
Training data & \multicolumn{5}{c}{SceneFlow + Asymmetric Augmentation} \\
\hline
PSMNet~\cite{chang2018pyramid} & 6.0 & 6.3 & 15.8 & 9.8 & 10.2 \\
GANet~\cite{zhang2019ga} & 5.5  & 6.0 & 13.5 & 8.5 & 6.5  \\
STTR \cite{li2021revisiting} & 8.7 & 6.7 & 15.5 & 9.7 & 17.2 \\
CFNet~\cite{shen2021cfnet} & 4.7 & 5.8 & 15.3 & 9.8 & \textbf{5.8} \\
FC-PSMNet (ours) & 5.3 & 5.8 & 15.1 & 9.3  & 9.5 \\
FC-GANet (ours) & \textbf{4.6} & \textbf{5.3}  & \textbf{10.2}  & \textbf{7.8}  & \textbf{5.8}  \\
\hline
\end{tabular}
\end{center}
\vspace{-6mm}
\caption{Cross-domain generalization evaluation on the KITTI, Middlebury, ETH3D training sets. Threshold error rates (\%) are adopted.}
\vspace{-2mm}
\label{table:exp:cross-domain_evaluations} 
\end{table}

\begin{table}[t]\footnotesize
\begin{center}
\begin{tabular}{c|c|cccc}
\hline
Methods & KITTI & Cloudy & Foggy & Rainy & Sunny \\
\hline
\hline
PSMNet \cite{chang2018pyramid} & 6.3 & 7.9 & 10.8 & 12.2 & 7.4 \\
FC-PSMNet (ours) & \textbf{5.8} & \textbf{4.3} & \textbf{6.2} & \textbf{7.2}  & \textbf{4.9} \\
\hline
GANet \cite{zhang2019ga}  & 6.0 & 5.7  & 8.2 & 10.0 & 5.4 \\
FC-GANet (ours) & \textbf{5.3}  & \textbf{3.3}  & \textbf{4.0} & \textbf{7.0} & \textbf{3.3}  \\
\hline
\end{tabular}
\end{center}
\vspace{-6mm}
\caption{Generalization evaluation on the KITTI and the half resolution DrivingStereo data sets of different weather conditions. Threshold error rates (\%) are adopted.}
\vspace{-3mm}
\label{table:exp:weather} 
\end{table}

\subsection{Fine-tuning on KITTI}
We evaluate the fine-tuned accuracy on the KITTI benchmark. The models are first trained on SceneFlow data and fine-tuned on the KITTI 2015 training set for a further 1000 epochs. During fine-tuning, we use the query encoder to extract two view features, which is the symmetrical feature extraction identical to the standard pipeline. We set the learning rate at 0.001 for the first 600 epochs and decrease it to 0.0001 for the rest 400 epochs. \Cref{table:exp:kitti_benchmark} shows results on the benchmark. We see that our models can obtain comparable performance to their counterparts. In addition, we explore the performance of our method with the more limited fine-tuning set, which is meant for practical applications because the data available for many real-world scenes is very limited. The KITTI (40) is a popular validation set \cite{chang2018pyramid}, which collects 40 images of representative scenes from the KITTI 2015 training set. And KITTI (1) only consists of the first training image. As shown in \Cref{table:exp:kitti_benchmark}, our models perform better than their counterparts with limited fine-tuning data.

\begin{table}[t]\footnotesize
\begin{center}
\begin{tabular}{c|ccc|ccc}
\hline
\multirow{2}{*}{Methods} & \multicolumn{3}{c|}{All-D1(\%)} & \multicolumn{3}{c}{Noc-D1(\%)}  \\ \cline{2-7}
& bg & fg & all &bg & fg & all \\
\hline
\hline
Fine-tuning set & \multicolumn{6}{c}{KITTI (full)} \\
\hline
PSMNet \cite{chang2018pyramid}  & 1.86 & 4.62 & \textbf{2.32} & 1.71 & 4.31 & 2.14 \\
FC-PSMNet (ours) & 1.86 & 4.61 & \textbf{2.32} & 1.73 & 4.19 & \textbf{2.13} \\
\hline
Fine-tuning set & \multicolumn{6}{c}{KITTI (40)} \\
\hline
PSMNet \cite{chang2018pyramid}  & 4.15  & 7.03  & 4.63  & 3.92  & 6.36  & 4.32  \\
FC-PSMNet (ours) & 3.10  & 6.94  & \textbf{3.74}  & 2.88 & 6.27  & \textbf{3.44}  \\
\hline
Fine-tuning set & \multicolumn{6}{c}{KITTI (1)} \\
\hline
PSMNet \cite{chang2018pyramid}  & 4.83  & 14.26  & 6.40  & 4.57  & 13.38	& 6.02  \\
FC-PSMNet (ours) & 3.34 & 12.56  & \textbf{4.87}  & 3.05 & 11.56  & \textbf{4.45}  \\
\hline
\end{tabular}
\end{center}
\vspace{-6mm}
\caption{After fine-tuning evaluation on the KITTI 2015 benchmark. Different subsets are used for fine-tuning.}
\vspace{-1mm}
\label{table:exp:kitti_benchmark} 
\end{table}

\section{Conclusion}
We have introduced a feature consistency idea to improve the domain generalization performance of end-to-end stereo networks. We propose to explicitly impose a contrastive loss on learned features during training to maintain the consistency between stereo views. Then we restrict the intermediate feature representations with a selective whitening loss, which helps to maintain the feature consistency on unseen domains. Experimental results show that our approach significantly improves the generalization performance of end-to-end stereo matching networks.

\section*{Acknowledgment}
This work was supported by the National Natural Science Foundation of China (Grant No.61772057 and 62106012), Beijing Natural Science Foundation (4202039).

\clearpage
{\small
\bibliographystyle{ieee_fullname}
\bibliography{egbib}
}

\clearpage
\def\thesection{\Alph{section}}
\setcounter{section}{0}
\appendix{{\textbf{\Large Appendix}}}


\section{Generalization Performance with the RGB Based Cost Volume} \label{appendix:rgb cost volume}
\begin{figure}[h]
\setlength{\abovecaptionskip}{0.cm}
\setlength{\belowcaptionskip}{-0.cm}
\centering
\begin{subfigure}{1\linewidth}
\centering
\includegraphics[width=1.0\linewidth]{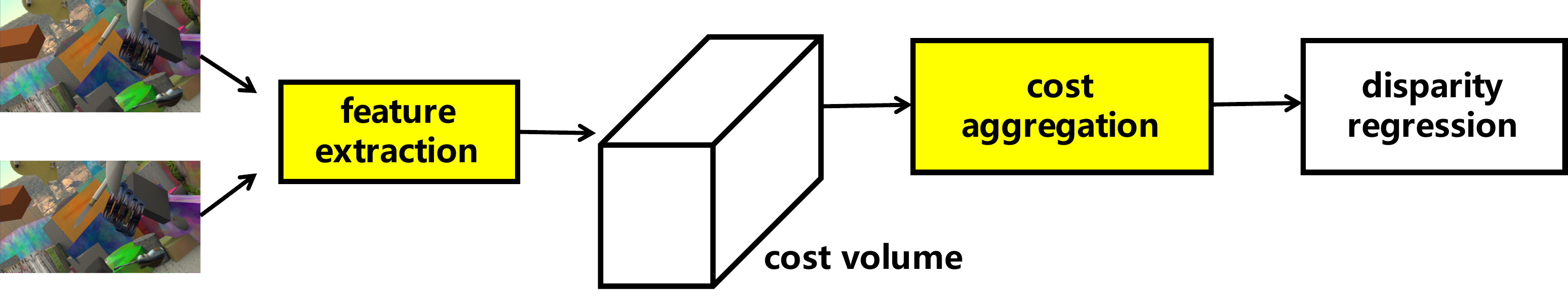}
\vspace{+1mm}
\label{fig:frame_rgbvolume(a)}
\end{subfigure}
\quad
\begin{subfigure}{1\linewidth}
\centering
\includegraphics[width=1.0\linewidth]{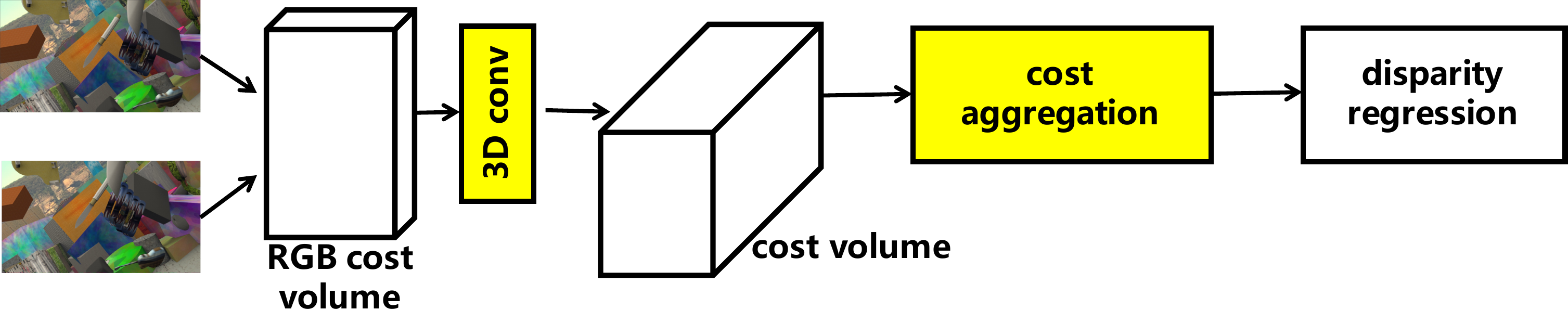}
\vspace{-1mm}
\label{fig:frame_rgbvolume(b)}
\end{subfigure}
\caption{The architecture of the original PSMNet with the concatenated feature cost volume (above) and its variant with the RGB based cost volume (below). Structures that contain learnable parameters are highlighted.}
\label{fig:frame_rgbvolume}
\end{figure}

\begin{figure}[h]
\centering
\includegraphics[width=0.8\linewidth]{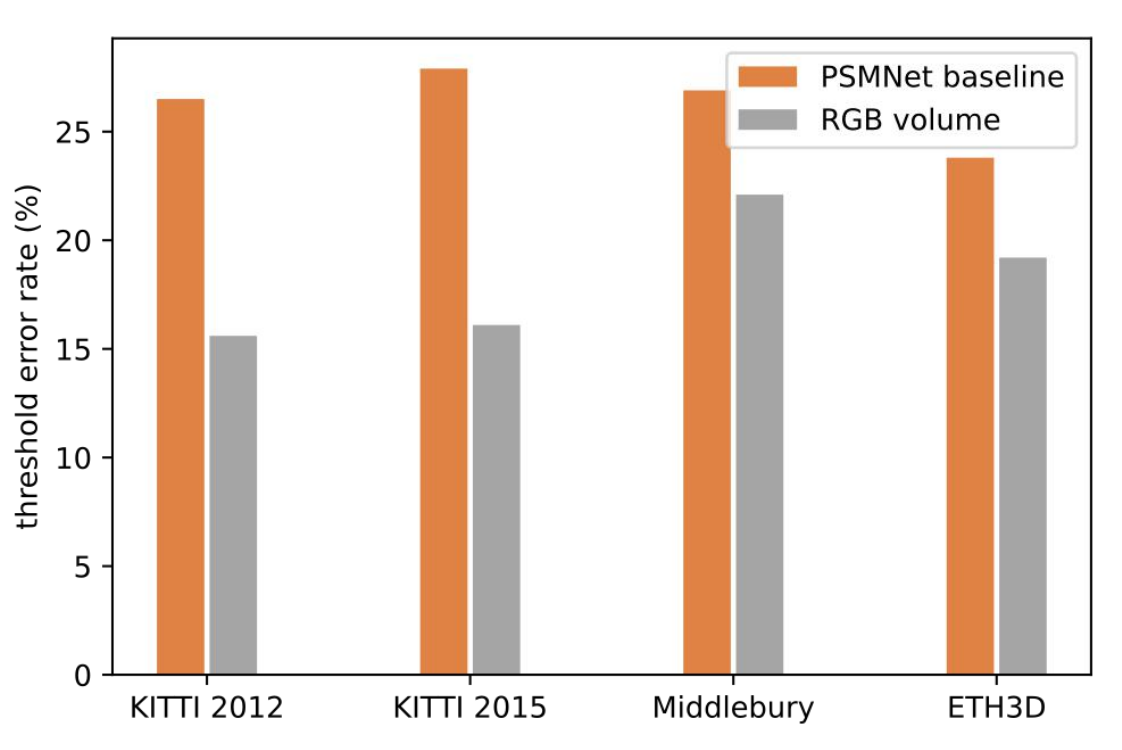}
\vspace{-3mm}
\caption{Generalization performance with cost volume construction using learned features and RGB images. Threshold error rates ($\%$) are used (KITTI: 3.0, Middlebury: 2.0, ETH3D: 1.0).}
\label{fig:RGB feauture}
\end{figure}

As mentioned in \Cref{sec:intro} of the main text, RGB images have consistent representations across stereo views, and traditional stereo matching methods generalize well across different domains. To further validate the importance of stereo representation consistency to generalization, we design a network that concatenates the RGB image pair directly to construct the cost volume, which is served as the input to the network. Compared to the PSMNet \cite{chang2018pyramid} baseline, we simply replace the concatenated feature-based cost volume with RGB based cost volume, as shown in \Cref{fig:frame_rgbvolume}. To match the input size requirement of the cost aggregation network ($D \times 64 \times \frac{H}{4} \times \frac{W}{4}$), a 3D convolution layer is applied to the RGB based cost volume ($D \times 6 \times H \times W$) that generates the cost volume as in PSMNet.  We eliminate as many learnable parameters from the feature extractor as possible and demonstrate that once the stereo feature consistency is highly satisfied, the training of the cost aggregation network doesn't affect the generalization performance of the whole stereo network. We present the generalization performance on different datasets in \Cref{fig:RGB feauture}. The error rate above a given threshold is used as the error metric. Constructing the cost volume directly on RGB image pairs provides a significant performance improvement over the cost volume with inconsistent stereo features (given by the feature extractor of PSMNet) on all datasets. This verifies our motivation to encourage stereo feature consistency for better generalization ability.

\vspace{-2mm}

\vspace{-5mm}
\section{Comparison with Pre-training method} 
The pre-training technique could obtain features that adapt to multiple domains and transfer to downstream tasks. For example, SAND \cite{spencer2019scale} utilizes metric learning to enforce feature consistency between stereo views, and uses these pre-trained features to fine-tune the stereo matching network. This pre-training method treats consistent feature learning and stereo network training as a two-stage framework. Nevertheless, our method jointly trains the stereo network and enforces the stereo feature consistency. We compare the feature consistency and the disparity error of the pre-training method \cite{spencer2019scale} and our joint training one in the whole training process. Our method produces more consistent features and lower disparity errors.

\vspace{-2mm}
\begin{figure}[h]
\centering
\includegraphics[width=1.0\linewidth]{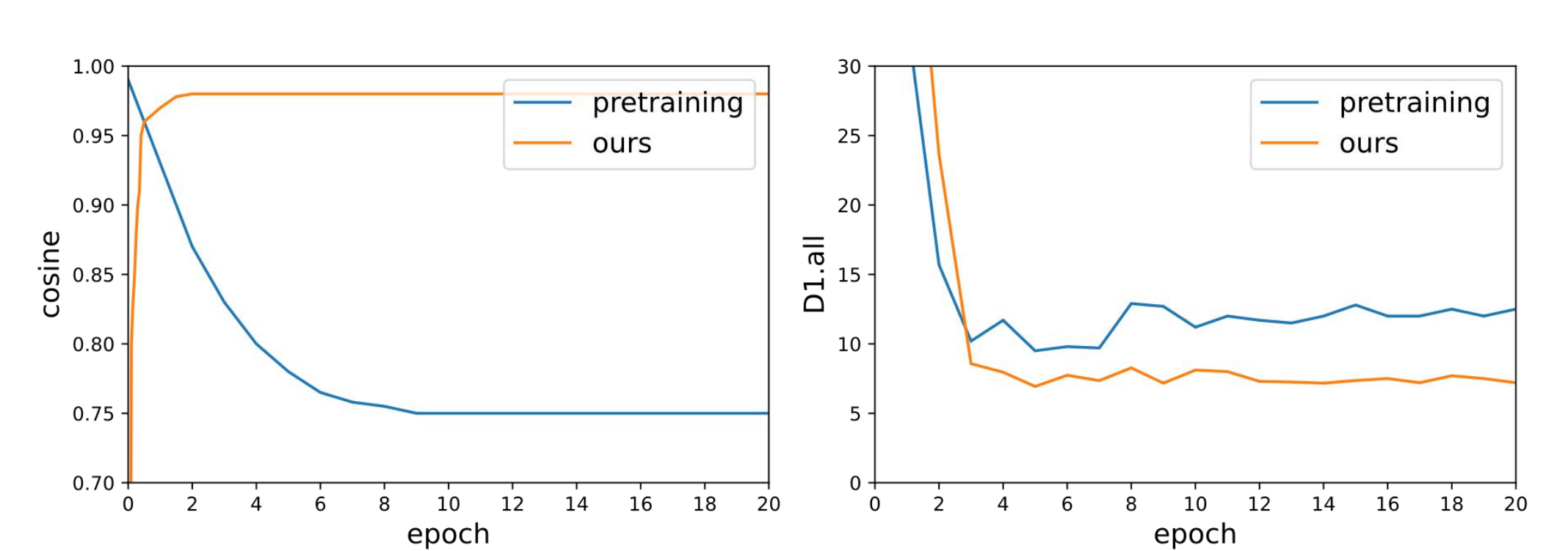}
\vspace{-5mm}
\caption{Comparison of enforcing feature consistency in a separate pre-training stage \cite{spencer2019scale} and jointly with stereo network training (ours). We evaluate both the stereo feature consistency and the generalization performance on the KITTI 2015 dataset. The cosine similarity and the D1\_all error rate are used to measure the stereo feature consistency and the disparity accuracy, respectively.}
\label{fig:pretrained features}
\end{figure}

\section{Channel Visualization of Feature Consistency} \label{appendix:feature consistency}
\begin{figure*}[h]
\centering
\includegraphics[width=0.88\linewidth]{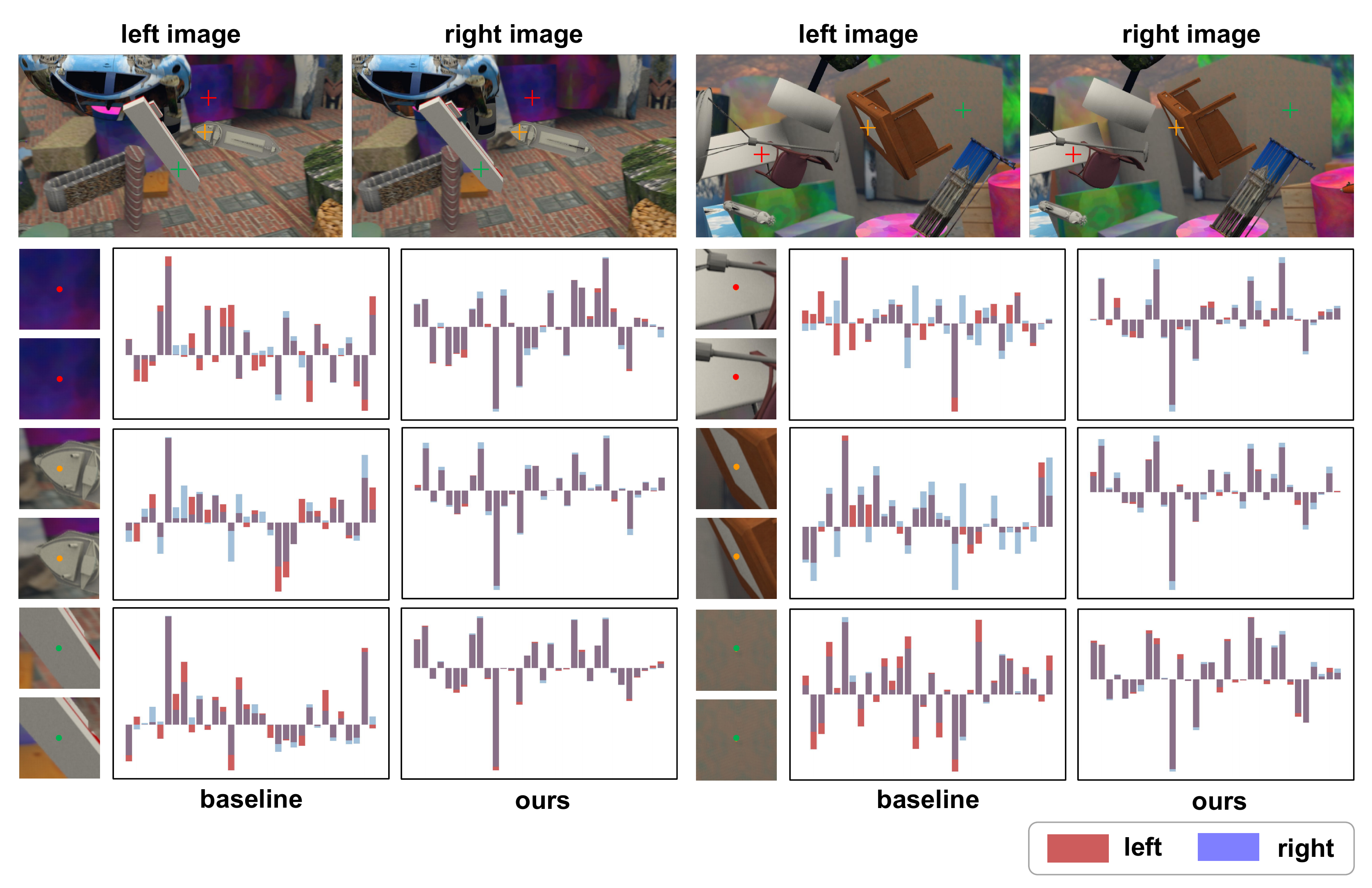}
\vspace{-3mm}
\caption{Channel-wise values of matching feature representations from the SceneFlow training set. Matching pixels are zoomed in to make them more visible.}
\label{fig:feature_detail_sf}
\end{figure*}

\begin{figure*}[h]
\centering
\includegraphics[width=0.88\linewidth]{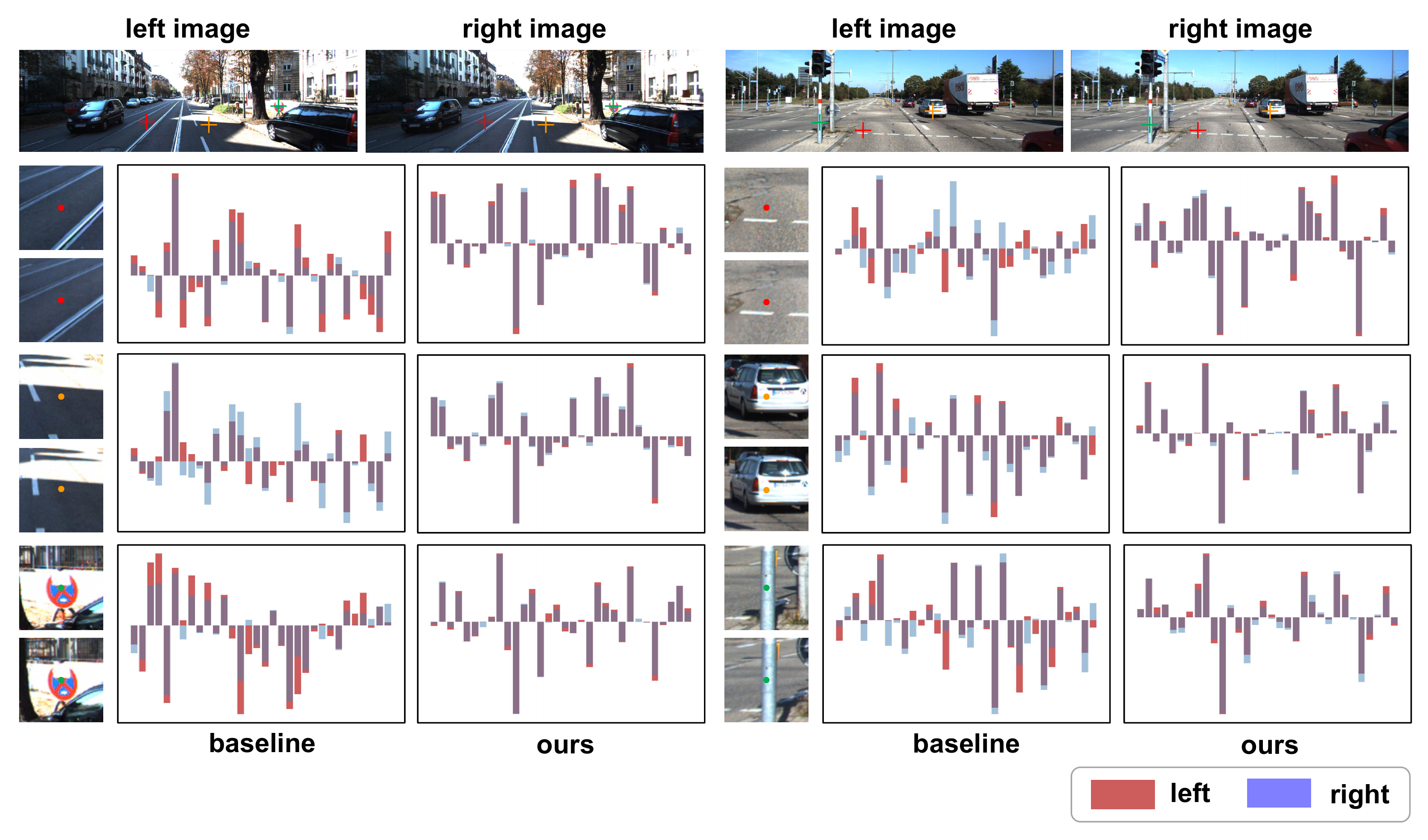}
\vspace{-3mm}
\caption{Channel-wise values of matching feature representations from the KITTI 2015 dataset. Matching pixels are zoomed in to make them more visible.}
\label{fig:feature_detail_kitti}
\end{figure*}

\begin{figure*}[h]
\centering
\includegraphics[width=0.88\linewidth]{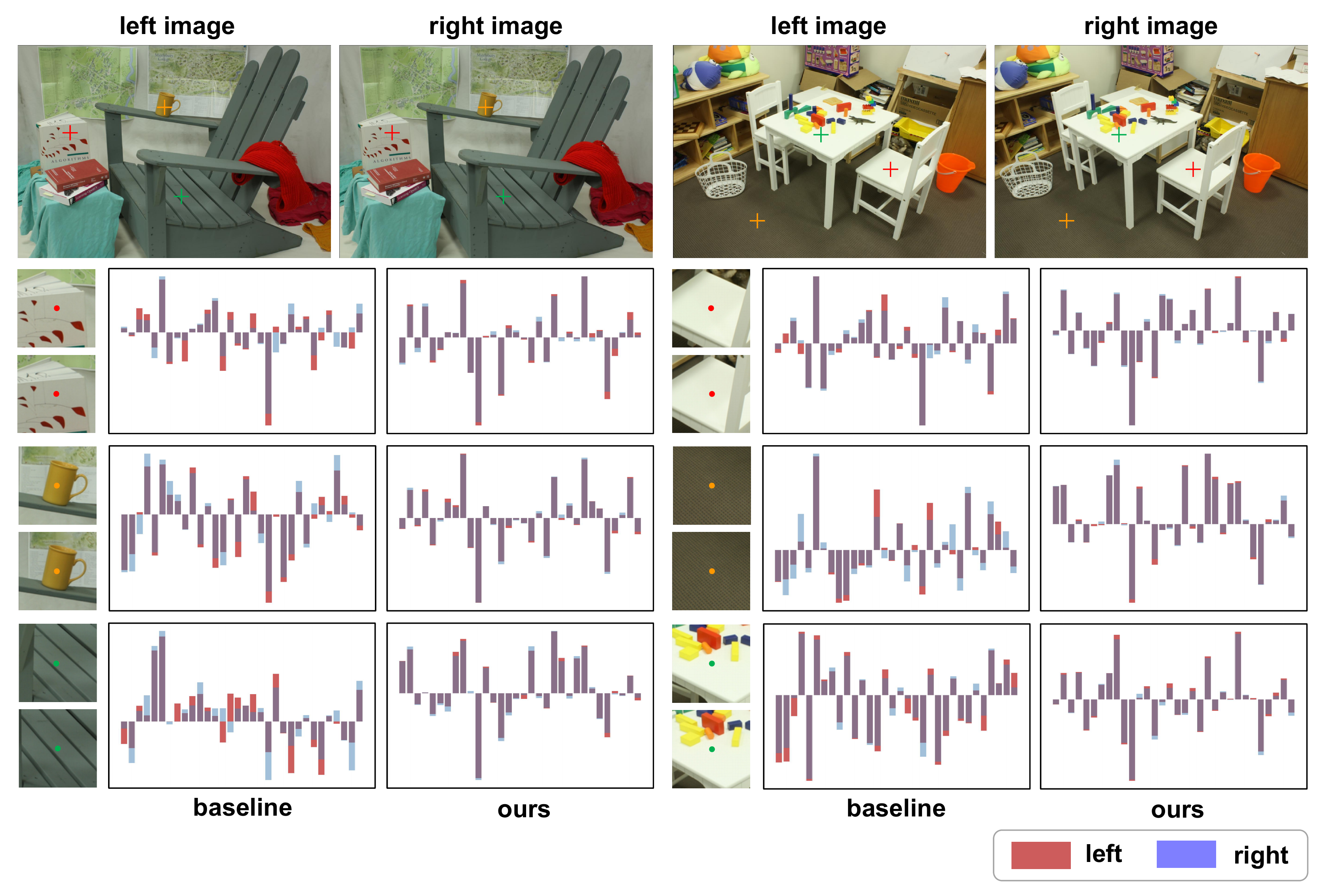}
\vspace{-3mm}
\caption{Channel-wise values of matching feature representations from the Middlebury dataset. Matching pixels are zoomed in to make them more visible.}
\label{fig:feature_detail_midd}
\end{figure*}

\begin{figure*}[h]
\centering
\includegraphics[width=0.88\linewidth]{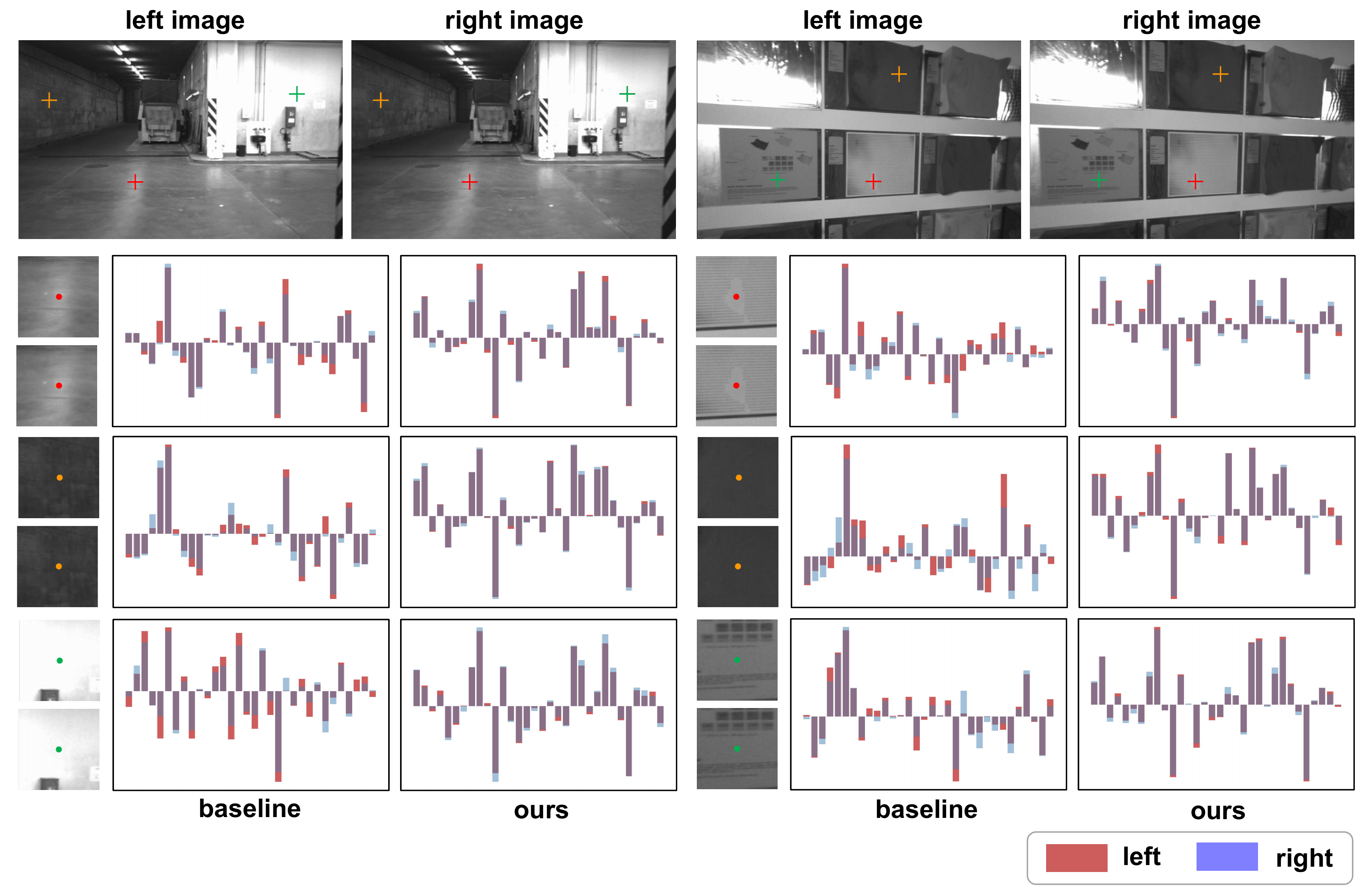}
\vspace{-2mm}
\caption{Channel-wise values of matching feature representations from the ETH3D dataset. Matching pixels are zoomed in to make them more visible.}
\label{fig:feature_detail_eth3d}
\end{figure*}

We visualize the feature vectors of some matching pixels and show that our method successfully improves the stereo feature consistency over the baseline method. Channel-wise values of left and right feature vectors are shown in \Cref{fig:feature_detail_sf,fig:feature_detail_kitti,fig:feature_detail_midd,fig:feature_detail_eth3d}, covering the training set and several unseen datasets. The PSMNet \cite{chang2018pyramid} baseline extracts feature representations that are inconsistent between stereo viewpoints, while our method can maintain the stereo feature consistency across different domains.

\section{More Qualitative Results}
\begin{figure*}[h]
\centering
\includegraphics[width=1.0\linewidth]{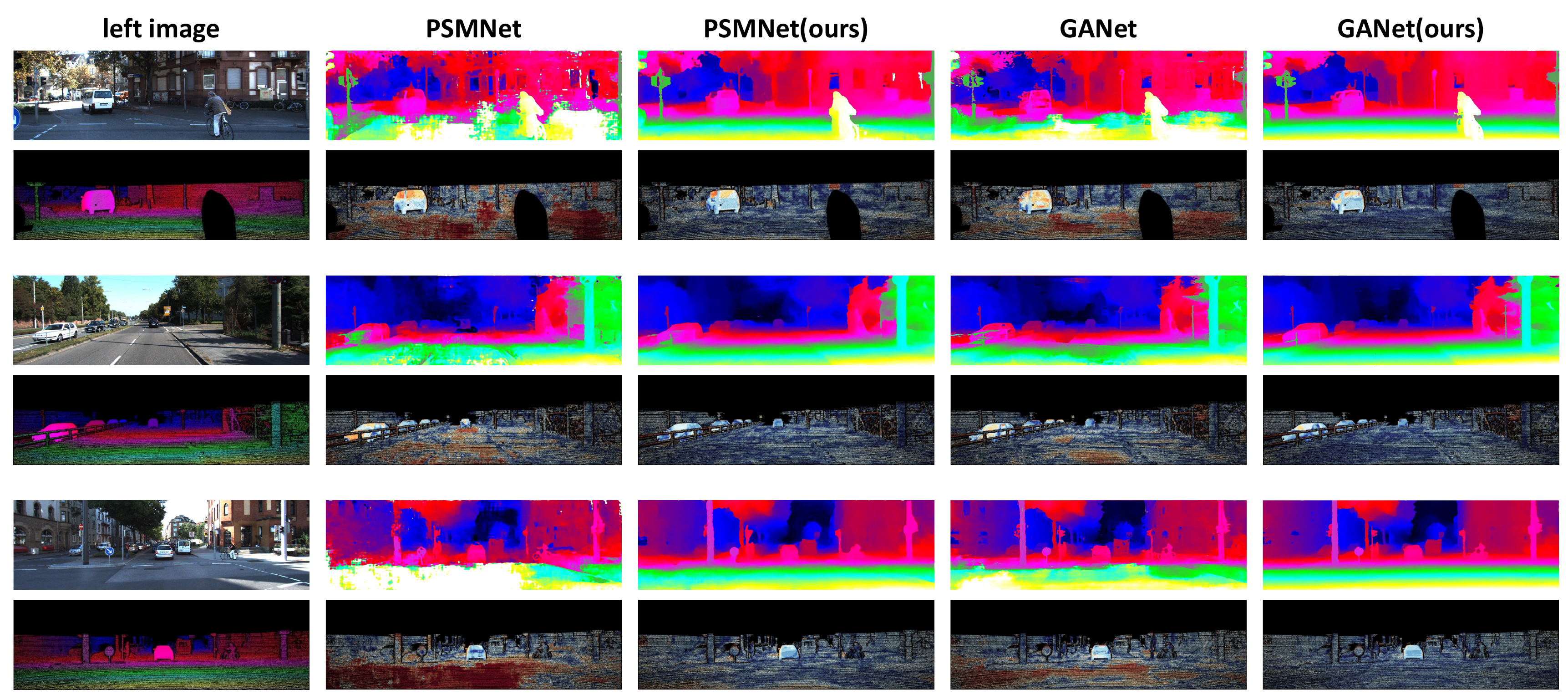}
\vspace{-5mm}
\caption{Qualitative results on the KITTI2015 training set. The left panel shows the left input image of the stereo pair and the ground truth disparity. And for each example, the first row shows the colorized disparity estimation and the second row shows the error map.}
\label{fig:more_results_kitti15}
\end{figure*}

\begin{figure*}[h]
\centering
\includegraphics[width=1.0\linewidth]{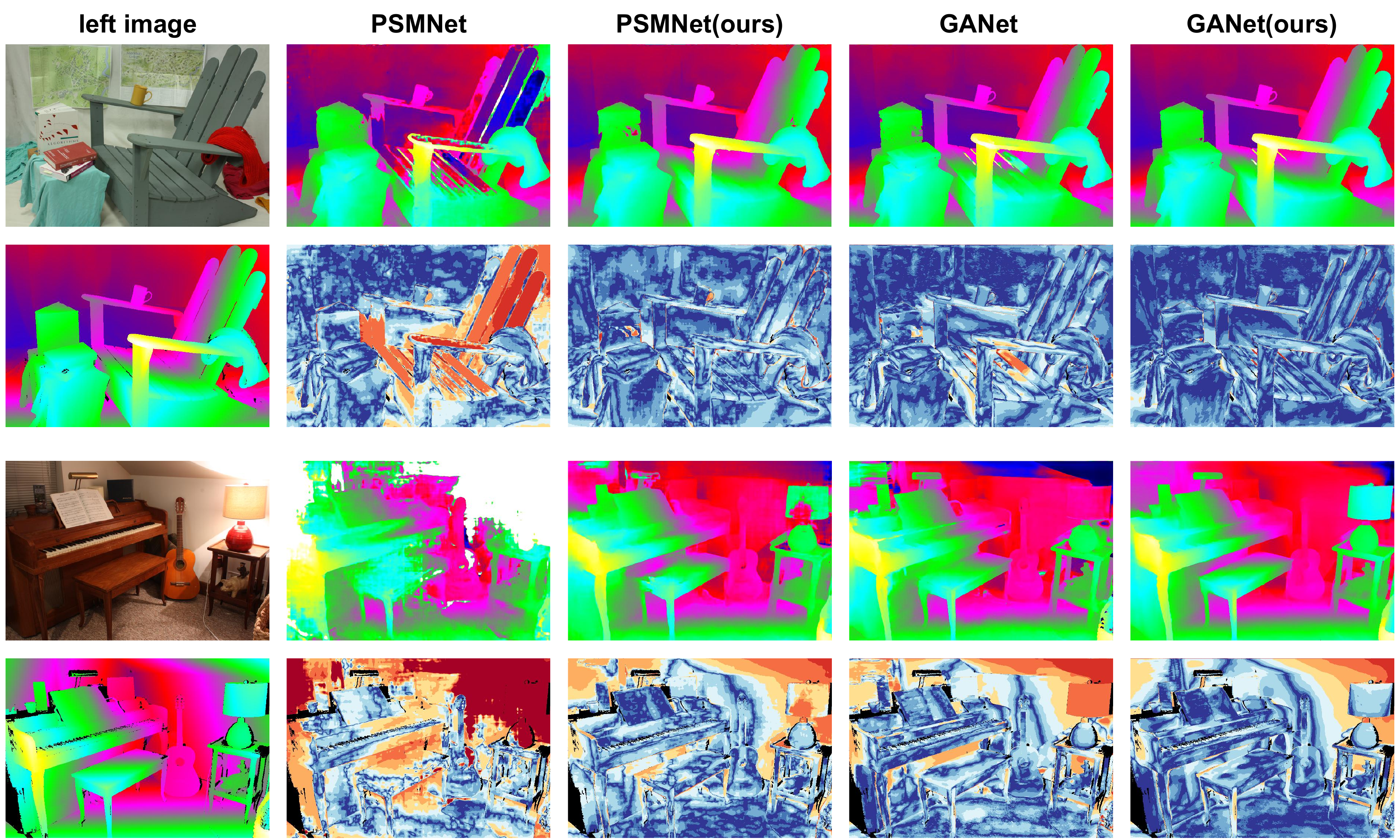}
\vspace{-5mm}
\caption{Qualitative results on the Middlebury training set. The left panel shows the left input image of the stereo pair and the ground truth disparity. And for each example, the first row shows the colorized disparity estimation and the second row shows the error map.}
\label{fig:more_results_midd}
\end{figure*}

\begin{figure*}[h]
\centering
\includegraphics[width=1.0\linewidth]{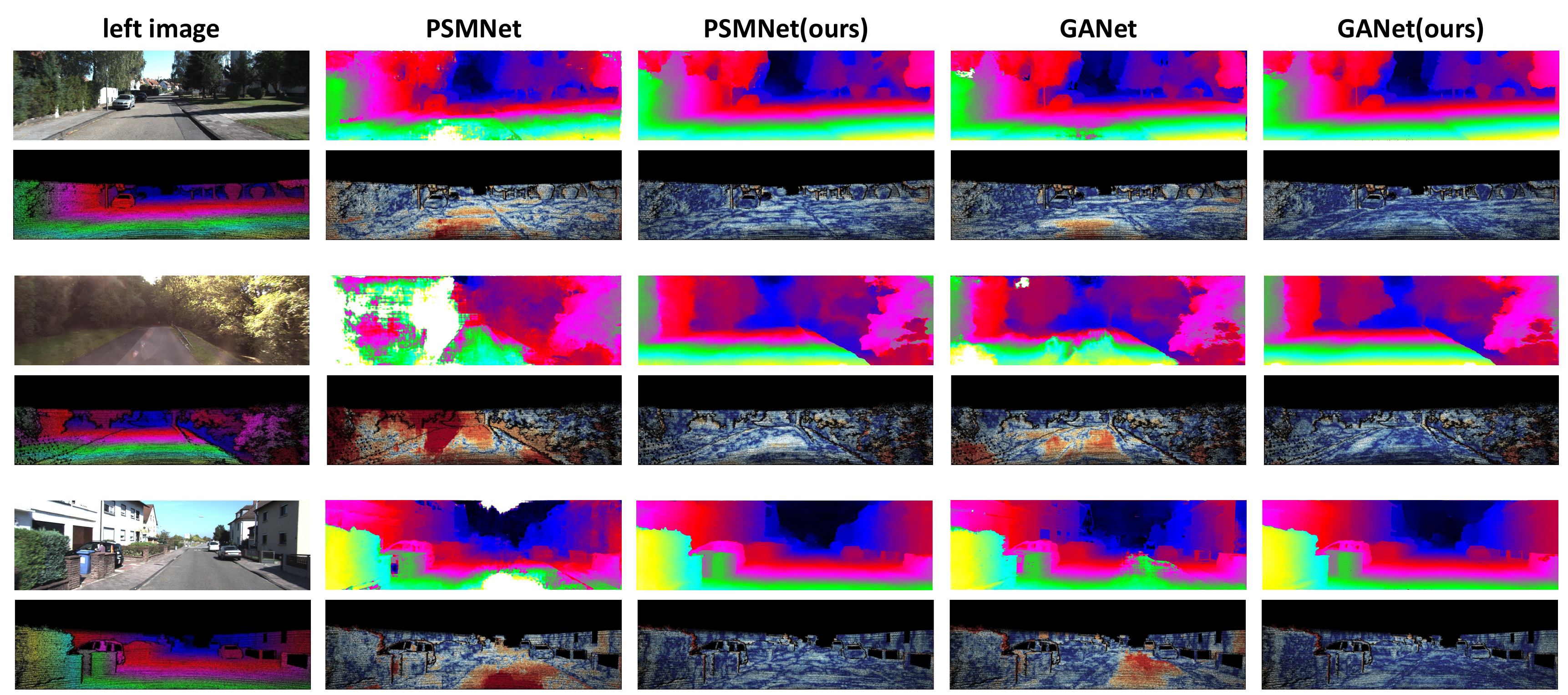}
\vspace{-3mm}
\caption{Qualitative results on the KITTI2012 training set. The left panel shows the left input image of the stereo pair and the ground truth disparity. And for each example, the first row shows the colorized disparity estimation and the second row shows the error map.}
\label{fig:more_results_kitti12}
\end{figure*}

\begin{figure*}[h]
\centering
\includegraphics[width=1.0\linewidth]{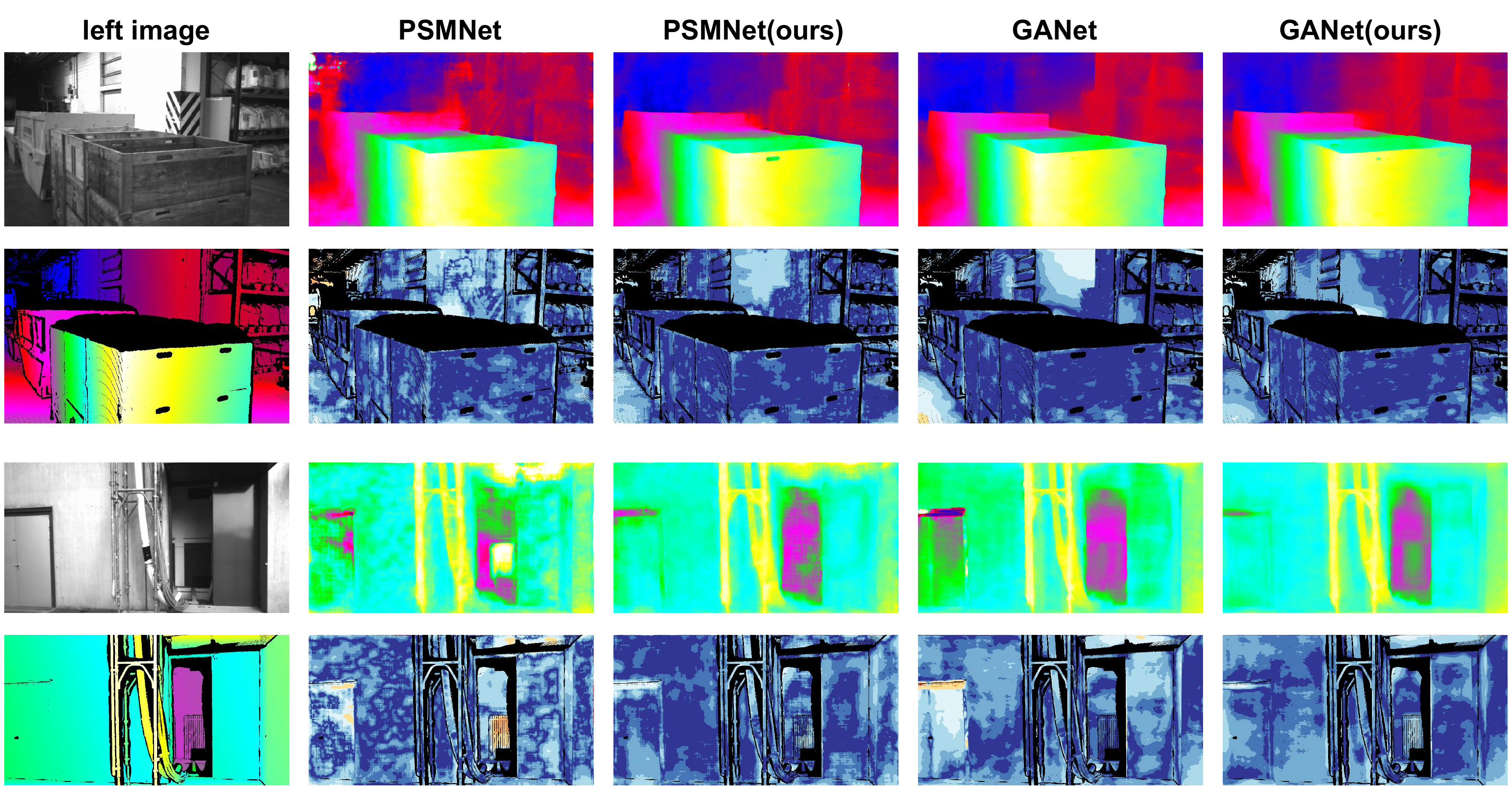}
\vspace{-3mm}
\caption{Qualitative results on the ETH3D training set. The left panel shows the left input image of the stereo pair and the ground truth disparity. And for each example, the first row shows the colorized disparity estimation and the second row shows the error map.}
\label{fig:more_results_eth3d}
\end{figure*}

In this section, we provide more qualitative results of baseline models and models with our method. PSMNet \cite{chang2018pyramid} and GANet \cite{zhang2019ga} are selected as our baseline models. All models are trained on the SceneFlow dataset and evaluated on four unseen domains. As shown in \Cref{fig:more_results_kitti15,fig:more_results_kitti12,fig:more_results_midd,fig:more_results_eth3d}, models with our method generalize better than their counterparts.


\newpage

\end{document}